\documentclass{article}

     \PassOptionsToPackage{numbers, compress}{natbib}

\usepackage[final]{neurips_data_2022}

\usepackage[utf8]{inputenc}     
\usepackage[T1]{fontenc}        
\usepackage{hyperref}           
\usepackage{url}                
\usepackage{booktabs}           
\usepackage{amsfonts}           
\usepackage{amsmath}            
\usepackage{nicefrac}           
\usepackage{microtype}          
\usepackage[dvipsnames]{xcolor} 
\usepackage{cleveref}
\usepackage{wrapfig}
\usepackage{sistyle}
\usepackage{graphicx}
\usepackage{array,multirow}
\usepackage{listings}
\usepackage{dialogue}

\newcommand*{\doi}[1]{\href{http://doi.org/#1}{doi: #1}}

\SIthousandsep{,}
\newcommand{\besto}[1]{\textbf{\textcolor{Periwinkle}{#1}}}
\newcommand{\bests}[1]{\textbf{#1}}

\title{\textsc{Fact}: Learning Governing Abstractions Behind Integer Sequences}

\author{
  Peter Belcak\thanks{The authors of this work are listed alphabetically.} \\
  ETH Zürich\\
  8092 Zürich, Switzerland\\
  \texttt{belcak@ethz.ch} \\
  \And
  Ard Kastrati \\
  ETH Zürich\\
  8092 Zürich, Switzerland\\
  \texttt{kard@ethz.ch} \\
  \AND
  Flavio Schenker \\
  ETH Zürich\\
  8092 Zürich, Switzerland\\
  \texttt{flaviosc@ethz.ch} \\
  \And
  Roger Wattenhofer \\
  ETH Zürich\\
  8092 Zürich, Switzerland\\
  \texttt{wattenhofer@ethz.ch} \\
}

\begin{document}

\maketitle

\begin{abstract}
    Integer sequences are of central importance to the modeling of concepts admitting complete finitary descriptions.
    We introduce a novel view on the learning of such concepts and lay down a set of benchmarking tasks aimed at conceptual understanding by machine learning models.
    These tasks indirectly assess model ability to abstract, and challenge them to reason both interpolatively and extrapolatively from the knowledge gained by observing representative examples.
    To further aid research in knowledge representation and reasoning, we present FACT, the Finitary Abstraction Comprehension Toolkit.
    The toolkit surrounds a large dataset of integer sequences comprising both organic and synthetic entries, a library for data pre-processing and generation, a set of model performance evaluation tools, and a collection of baseline model implementations, enabling the making of the future advancements with ease.
\end{abstract}

\section{Introduction}
\label{section:introduction}
Ordered lists of integers are the natural representation form for all fundamentally discrete abstractions.
These arise when encountering evolutions of discrete-time phenomena, finite symmetries of visual patterns, or algorithmic progressions, where they describe the development of consecutive states of a system, automorphisms of $\mathbb{R}^2$, or program listings, respectively.
Sequences of integers are also the representation of choice when linearising structured information for analysis, data compression, and communication, with often-appearing datapoints tending to be encoded in the simplest or shortest form.
Testifying to their utility to accurately represent abstractions, completion and extrapolation tasks on integer sequences are a frequent part of general human intelligence and aptitude testing (\cite{strannegaard2013anthropomorphic, russell2003times}).

It is the aim and the hope for machine learning models to identify straightforward universal abstractions explaining the training data, rather than to memorise a plethora of small classes of exemplars and interpolate when given previously unseen input.
The discovery and internalisation of governing concepts, or simply the learning of underlying rules, thus sits at the centre of artificial intelligence research.

We note that many concepts may be uniquely represented by a sequence of integers naturally (e.g. the squares of the natural numbers determine the polynomial $n^2$; $123,312,231,132,321,213$ encodes the symmetries of a triangle), while others (such as the rotations of objects in scene) require a continuous space for a proper, scalable description. 
We recognise integer sequences as a general form for description of concepts completely representable with finite precision (finite in their nature; \textit{finitary}) and put them at the centre of our study.

\begin{figure}
    \centering
    \includegraphics[width=1\columnwidth]{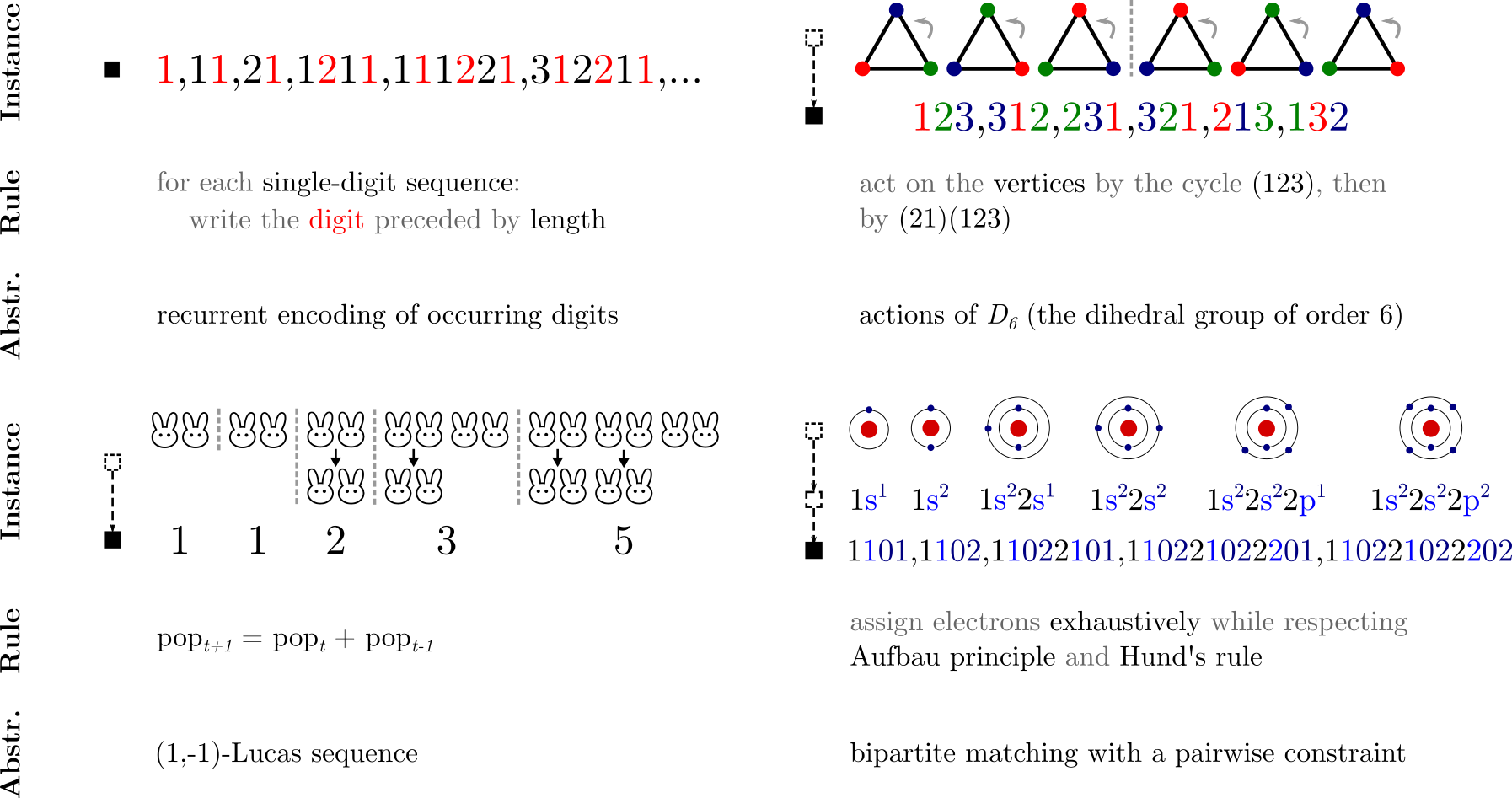}
    \caption{
        An illustration of an example conceptual learning programme applied to four separate instances: a toy recursive digit-counting sequence, the symmetries of an equilateral triangle, the evolution of an idealised rabbit population as described in \cite{sigler2003fibonacci}, and the electron configurations across shells and sub-shells of an atom.
        In each instance, the raw data is comprehended in its original modality and then turned into an integer sequence.
        Then, a prior for the rule that applies is formulated.
        As more samples are observed, the rule is iteratively generalised until the governing abstraction has been fully revealed.
    }
    \label{fig:learning-programme}
\end{figure}

To discern the learning and understanding of these discrete abstractions from the learning of their representations in various modalities, we introduce a rich dataset of integer sequences together with a compendium of corresponding tasks that are innately related to integer sequences and well-suited to assess the levels of human-like understanding exhibited by machine learning models.
Focusing solely on integer sequences, we thus set the concepts being learned apart from virtually all complexity stemming from the learning of a representation, making the learning process less resource-intensive and the interpretation of evaluation results more straightforward.
An example learning programme making this distinction and proceeding to high-level abstractions is pictured in \Cref{fig:learning-programme}.

Modern machine learning methods have been shown to posses the ability to comprehend (or at least pattern-match) convoluted concepts appearing in various data modalities, especially by the means of using deep learning to construct informative representations of the entities studied.
In spite of working with number sequences, we take a step away from symbolic regression (which has so far dominated the notion of \textit{understanding} in the area) and, tending to the trend, employ instead a multi-faceted approach in which sequences are characterised by their properties and relations to other sequences, rather than by explanatory symbolic formulas that are more readily interpretable by humans. We expand on the relationship of our work to symbolic regression in \Cref{appendix:relationship-to-symbolic-regression}.

Aiming at the comprehension of abstractions behind concrete representations of finitary phenomena, we unlock a new mode for evaluation of the quality of the abstractions learned.
A guess, or an estimate, of the rule behind an integer sequence, which further leads to correct predictions of the sequence's elements on previously unseen inputs, is arguably more desirable than an estimate that describes the sequence well only for inputs known in training.
The fundamentally algorithmic nature of the problem of learning finitary abstractions thus makes the problem of extrapolative generalisation well-defined, and allows us to consider extrapolative generalisation performance as a criterion for model assessment.

Our contributions are:
\begin{itemize}
    \item the introduction of a large dataset of integer sequences comprising data from both organic and synthetic sources and curated for subsequent use in tasks challenging models to develop understanding of the concepts determining the data (\Cref{section:dataset}, \cite{fact_dataset_through_doi}),
    \item complementing the above, a utility library (FACTLIB \cite{factlib_through_doi}) for integer sequence data processing and generation,
    \item the introduction of a variety of tasks designed to evaluate the model comprehension of conceptual patterns in integer sequences with a clearly established order of difficulty (\Cref{section:benchmark}),
    \item a battery of evaluation metrics tailored to the above tasks to appropriately assess model performance and track progress in this sub-area of knowledge representation and reasoning, and
    \item a collection of baseline models, both classical and neural, implemented to facilitate seamless experimentation (\Cref{section:baselines}, \cite{fact_benchmarking_baselines_through_doi}).
\end{itemize}

\section{Dataset}
\label{section:dataset}
As a part of FACT, we introduce a dataset consisting of over $3.6$ million integer sequences.
The structure-giving starting point for the dataset was the data made available by the Online Encyclopedia of Integer Sequences (\cite{sloane2007line}).
The OEIS is an organically grown comprehensive reference on noteworthy sequences of integers, compiled over decades to aid work in mathematical sciences.
We have reviewed the OEIS4 data, set apart a suitable subset of 341,000 entries, and processed it specifically for use in machine learning, in line with the license requirements.
Each entry of the dataset is now annotated by up to 18 features conveying the information about the nature, properties, and purpose of the sequence.
In \Cref{fig:hierarchy}, we give an overview of the result of this processing step.
A full discourse on the extensive curation, refining, and automated annotation effort undertaken can be found in \Cref{appendix:oeis}. 

\begin{figure}
    \centering
    \scalebox{1}{
        \includegraphics[width=0.75\columnwidth]{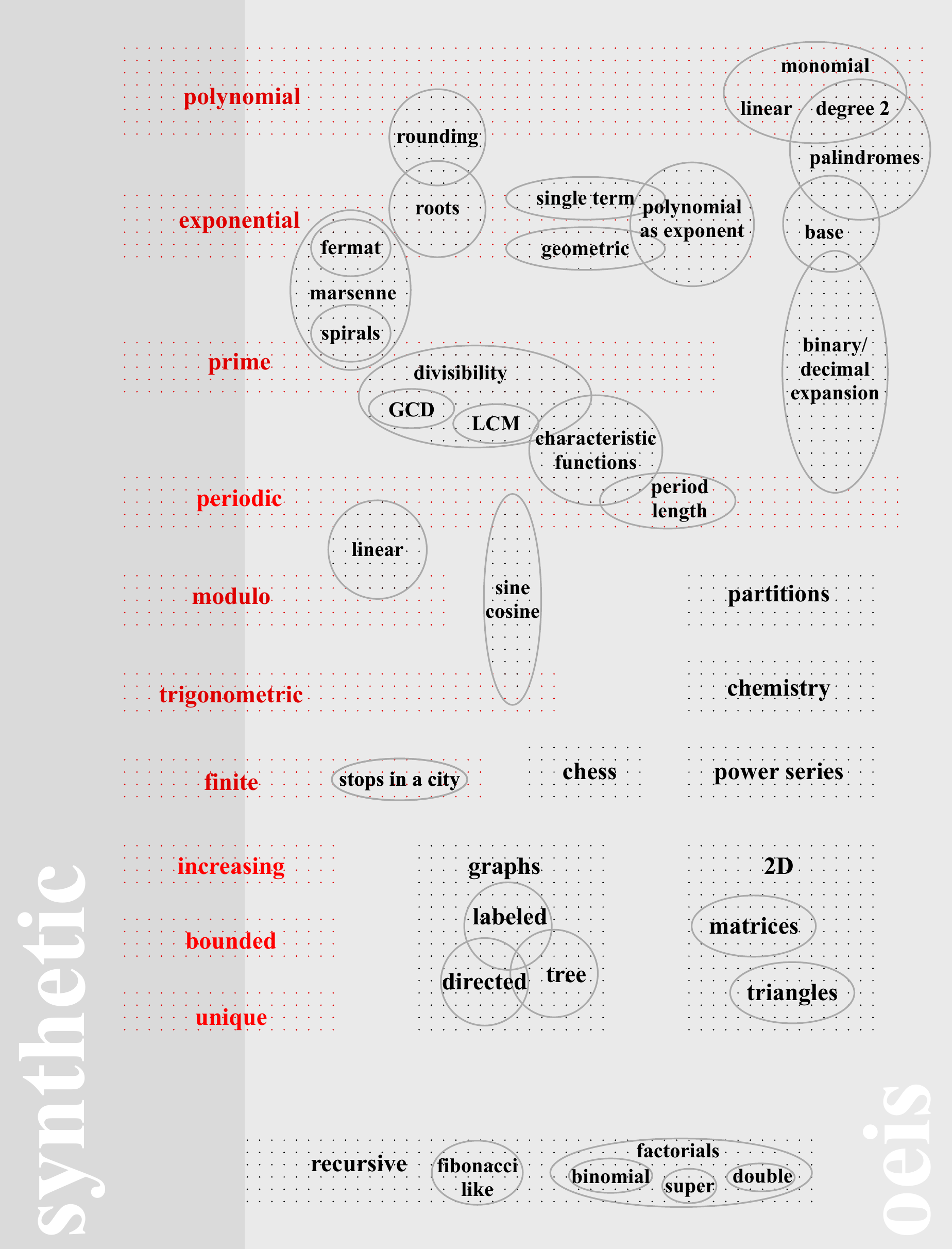}
    }
    \caption{
        The categories in the FACT dataset.
        It is composed of synthetic and OEIS entries.
        Each group in the synthetic part consists of 500,000 sequences, whereas the sizes of the OEIS groups vary.
        Dotted regions represent the main categories identified in the dataset.
        Ellipses define the sub-categories from our processing step in OEIS.
        Red dots mark groups that are augmented with synthetic data (and used in our benchmarking setup).
    }
    \label{fig:dataset}
    
\end{figure}

With the encyclopedia entries aimed at a human reader, we observed that many covered their respective categories only very sparsely, relying on the associated natural language descriptions and the human ability to abstract to make the categorical connection.
Our initial experiments with the baseline models (cf. \Cref{section:baselines}) further confirmed that much of the data did not reach the critical mass of information necessary for reliable use in machine learning applications.
We hence systematically extended the dataset by synthetically generated sequence branches while abiding by the structure and nature of the stem encyclopedia entries and providing carefully engineered automatic annotations wherever possible.

\subsection{Synthetic Generation}

\begin{wrapfigure}[16]{r}{0.57\linewidth}
    \centering
    \includegraphics[width=0.57\columnwidth]{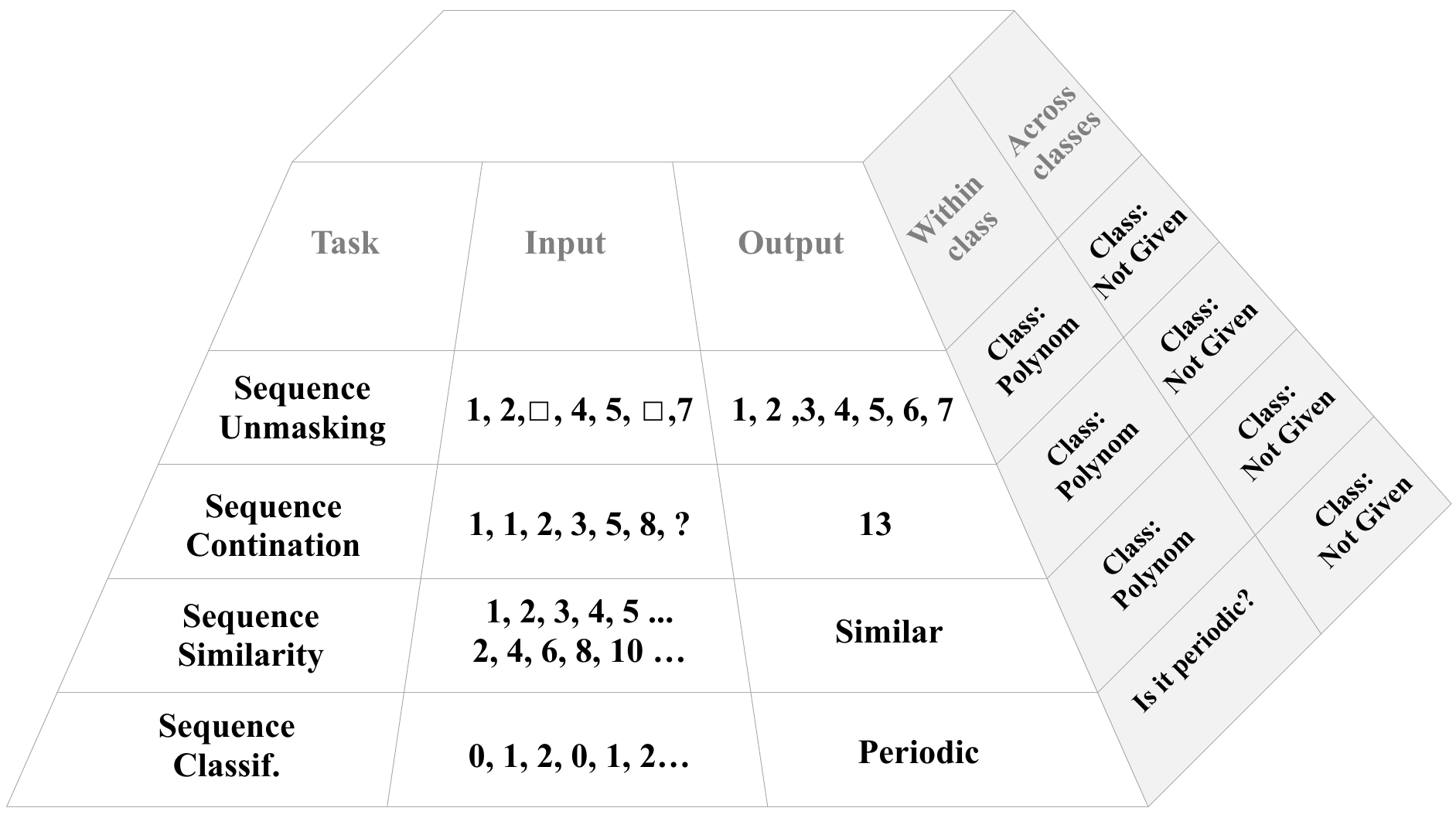}
    \caption{
        Our tasks ordered by the level of difficulty over two dimensions: type and scope.
    }
    \label{fig:hierarchy}
\end{wrapfigure}

Our principal inspirations were the notions of Kolmogorov complexity and Solomonoff probability \cite{schmidhuber, kolmogorov, solomonoff}.
Starting on the level of categories (cf. \Cref{fig:hierarchy}), we defined a context-free grammar $\mathcal{G}_c$ for each category $c$, and then used $\mathcal{G}_c$ to generate ever longer formulas, which were in turn used to generate the sequences.
This was done abiding by the notion that the growing length of formulas reflects itself in increasing complexity of rules and therefore generated sequences.
An example grammar, used for the production of polynomial formulas, is given in Figure \ref{fig:polynomials}.

For each category, a total of 500K synthetic sequences have been generated. 
The length of the formulas used to generate these sequences was continuously increased following a logarithmic schedule, thereby favouring shorter formulas while still ensuring the presence of sequences from longer formulas.
We give all details of our generation procedure for each sequence category in \Cref{appendix:synthetic}.

\begin{figure}
    \centering
\begin{lstlisting}
    T -> Var | Const
    N -> Add | Sub | Mult | Pow | NConst | T
    Add -> (N + N)
    Sub -> (N - N)
    Mult -> (N * N)
    Pow -> (N ** NConst)
    NConst -> (ConstPos NConst) | Const
    Var -> x
    Const -> 0 | 1 | 2 | 3 | 4 | 5 | 6 | 7 | 8 | 9
    ConstPos -> 1 | 2 | 3 | 4 | 5 | 6 | 7 | 8 | 9
\end{lstlisting}
    \caption{The context-free grammar used for the synthesis of formulas used for the generation of Polynomial sequences. $NConst$ denotes a number constant, $T$ denotes a term, and $N$ denotes the root non-terminal for the polynomial expression.}
    \label{fig:polynomials}
\end{figure}

The combined result of the OEIS curation and the dataset extension process is therefore a large dataset seamlessly integrated into FACT and easily extensible by FACTLIB, if required by more complex tasks or larger applications.

\section{Benchmark}
\label{section:benchmark}
Based on the dataset of the previous section, we propose a set of benchmarking tasks and evaluation metrics to assess a wide range of methods for their understanding of governing concepts behind evolutions of integer sequences.



\subsection{Motivation}
\label{section:motivation}
Consider the initial segment $0,2,4,6,8,\dots$ of a sequence and assume we are tasked with proposing reasonable candidates for the number that follows $8$.

How could we go about evaluating the quality of our suggestions?

In the spirit of symbolic regression, we may choose to insist that there must be a single formula that produces the members of the sequence in order.
But, even under such condition, for every suggested continuation integer there exists a degree-$8$ polynomial accommodating the continued sequence.
This is despite the fact that a human would intuitively be most likely to suggest $10$ as a reasonable continuation, perhaps even justifying it by the observation that the first $5$ elements of the sequence follow the pattern of $(2n)_{n \geq 0}$.

As described in \Cref{section:dataset}, we have therefore set the generation methods of the synthetic part of our dataset to explicitly apply the parsimony principle by varying the length of generative formulas to control the complexity of the resulting sequences.
Data generated in this manner is guaranteed to encompass sequences coming from rules of varying degrees of complexity by design, rather than by chance.

The focus on preferring shorter rules over longer ones would, however, be too artificial if employed alone.
Consider the initial segment $1,1,2,3,5,\dots$.
While it is tempting to promptly claim that the numbers come from the Fibonacci sequence and that $8$ should follow, an answer arising more naturally in the context of chemistry is $9$, as the continued sequence represents the count of all possible $n$-carbon alkanes.

Hence, striding ahead of the symbolic regression under Occam's razor, we add a processed part of the OEIS dataset into our evaluation procedures as an indicator of the sequences' appeal across a multitude of scientific disciplines.
Note that this is an improvement made in addition rather than in contrast to the principles for synthetic sequence generation, as many of the real-world sequences collected in OEIS do indeed obey straightforward rules.

\subsection{Structure}
As motivated in \Cref{section:motivation}, for each benchmarking task we provide two evaluation sets: one for synthetic data and one for OEIS data.
Our benchmark thus pushes for the design of machine learning models that identify simple concepts and rules (evaluation on synthetic data) but still retains enough of cross-domain generality to have practical impact in different disciplines (evaluation on OEIS).

The benchmark consists of tasks with an established order of difficulty over two dimensions: the task type and scope.
We distinguish between the tasks of sequence classification, sequence similarity, next sequence part prediction, sequence continuation, and sequence unmasking -- each detailed below.

In addition, we perform each task to the extent of two different scopes: within and across categories.
The case of performing the task within categories is the simpler of the two setups, since the category from which the sequence originates is known in advance, and this information is thus also available at the time models are designed.
Nevertheless, our baselines (cf. \Cref{section:baselines}) are oblivious to this information and instead treat the category scope as if nothing was known about the data.

In our particular benchmarking setup, our dataset is split into four parts, namely the training, validation, synthetic test, and organic test sets.
The training and validation sets consist of only synthetic sequences.
The size ratios between training, validation, synthetic testing, and OEIS testing datasets are 9:1:1:1.
For convenience and future reference, we make this divided-up data available as separate datasets.
Nevertheless, the FACT dataset is also available in one piece, allowing users to choose their own splits or supplement their own data generated by FACTLIB, according to their needs.

\subsection{Task Types}
\label{section:task-types}
In this section we present 5 types of tasks in order of difficulty, for which we take the upper bound on the difficulty of the task instances comprising the given task.
For example: Every instance of sequence continuation belongs to the set of sequence unmasking instances.
But, for every instance of sequence continuation (except for the trivial one where we only begin with one number) there is an instance of unmasking that is harder. Such an instance can be formed by further asking to unmask a sequence element somewhere in the initial segment provided for continuation.
Hence, the upper bound on the difficulty of sequence unmasking is strictly higher than the upper bound on the difficulty of sequence continuation.

\subsubsection{Sequence Classification}
The simplest task type in our benchmark is classifying in which category the sequence belongs.
The chief goal of this task is to evaluate whether models can distinguish and identify general patterns in sequences, both across and within different categories.
Note that category membership may not necessarily be unique.
For example, a sequence can be bounded, but also periodic.
For this task we use all categories that possess a synthetic counterpart within our dataset.
Each sample consists of an array of integer numbers for the given sequence class.
We distinguish between two task sub-types and give their objectives:
\begin{itemize}
    \item \textbf{One-vs-Rest (OvR).} \textit{Obj.} to identify whether the sequence belongs to a specified category or not. For this case, we provide a balanced dataset in each category. This is a binary classification task, and as such, we use accuracy as the evaluation metric. 
    \item \textbf{Multiclass classification.} \textit{Obj.} to predict, for every sequence, all categories to which it belongs. The performance is measured with the macro average F1 score (i.e. the mean of individual per-class F1 scores) due to inherent imbalances in our dataset.
\end{itemize}

\subsubsection{Sequence Similarity}
The similarity task aims to assess model ability to represent sequences in a way that reflects their similarities in type (e.g. agreeing on category membership) or properties (such as being periodic or unbounded) in the spirit of \cite{mikolov2013efficient}.
The objective is to embed sequences into an embedding space such that sequences belonging to the same category are closer to one another than to sequences of categories to whom they do not belong.
We evaluate using
\begin{itemize}
    \item the Recall@$k$ score, where the $k$ candidates for a sequence $s$ are proposed by sampling $k$ sequences from each category and then ordering the category labels according to the distance of the carrier points from $s$; and
    \item the top-$k$ root mean squared error -- the root mean squared error (RMSE) across the top $k$ similarity candidates.
    Given $k$ predictions of a model $\{\hat{y}_i\}_{i \in \{1,...,k\}}$ and ground truth $y$, we define the top-$k$ RMSE as $ \min_{i \in \{1,...,k\}} \text{RMSE}(y,\hat{y}_i)$. In other words, given all generated predictions we report the RMSE of the prediction closest to the ground truth.
\end{itemize}
Our choice of evaluation metrics is grounded in the observation that sequences generated from similar simple rules often eventually diverge, and that to a large extent.
This task generalises sequence classification.

\subsubsection{Next Sequence-Part Prediction}
As seen in natural language processing \cite{devlin2018bert}, asking a model to decide whether two sentences follow one another can be beneficial for testing of whether a model understands its inputs.
Given two contiguous sub-sequences $s_1$ and $s_2$, the objective of the next sequence-part prediction (NSPP) task is to determine whether the sub-sequence $s_2$ is a valid continuation of $s_1$. We create a balanced dataset for this task, by using all categories of our dataset that have a synthetic counterpart. 
NSPP is then simply a binary classification task and the performance is measured by prediction accuracy.
This task is strictly more difficult than the similarity task, since it demands that the model not only understands the key properties of the sequence but also possesses the ability to discern unlikely or unfeasible combinations of sequence parts from the feasible ones.

\subsubsection{Sequence Continuation}
The fourth task type in our difficulty hierarchy is to suggest the next entry in a given sequence $s$.
This task is \textit{extrapolative} in its nature, and is meant to challenge model understanding beyond making a binary decision between externally provided suggestions. As such, this task demands better understanding of the rules governing sequences than next sequence-part prediction -- hence it's placement above NSPP in the difficulty hierarchy.
We distinguish two sub-types of this task, namely the single-shot and multi-shot continuations, differing only in the number of candidates the model is expected to provide for the continuation.
We use the root mean squared logarithmic error (RMSLE) and top-$k$ RMSE for each of the sub-types, respectively.

\subsubsection{Sequence Element Unmasking}
At the apex of our complexity hierarchy is the task of unmasking marked elements of a provided sequence.
The sequence continuation task can be viewed as a special case of unmasking, where only the last element is masked.
We consider only multi-shot unmasking and choose top-$k$ RMSE as the evaluation metric.

\section{Baseline model performance}
\label{section:baselines}
We run extensive experiments on the proposed benchmark as a starting point for further research. We consider classical machine learning methods as well as large neural networks. The experiments in this section highlight the feasibility of learning many different patterns in integer sequences, but also some of the limitations of the existing methods.  

\subsection{Models}
To provide baselines for model performance on the above tasks, we use a total of 24 different models across our benchmarking tasks, namely 4 neural models (dense, recurrent, and convolutional networks, and transformers), 9 classical classifiers ($k$-nearest neighbours, Gaussian naive Bayes, linear support vector machine, decision tree, random forest, gradient boosting, AdaBoost, XG Boost, dummy classifier), and 11 standard regressors ($k$-nearest neighbours, linear regressor, ridge regressor, lasso regressor, Elastic Net, single decision tree, random forest, gradient boosting, AdaBoost, dummy regressor).

We give details on their implementations and hyperparameter settings in \Cref{appendix:baselines}.

\subsection{Results}
A simple overview of the performance of the baseline models can be found in \Cref{table:main-results}.
A comprehensive, detailed listing of our results, including results of evaluations on the category level, can be found in \Cref{appendix:baseline-performances}.

\begin{table*}[h!]
\centering
\scalebox{0.75}{
    \begin{tabular}{l|c|ccccc}
    
    \toprule
    Model & Dataset & \multicolumn{5}{c}{Task}\\
    \midrule
    
    
    \multicolumn{2}{c}{} &
    \parbox[t]{12mm}{\multirow{4}{*}{\centering\rotatebox[origin=c]{60}{classification}}} &
    \parbox[t]{12mm}{\multirow{4}{*}{\centering\rotatebox[origin=c]{60}{next part pred.}}} &
    \parbox[t]{12mm}{\multirow{4}{*}{\centering\rotatebox[origin=c]{60}{continuation}}} &
    \parbox[t]{12mm}{\multirow{4}{*}{\centering\rotatebox[origin=c]{60}{similarity}}} &
    \parbox[t]{12mm}{\multirow{4}{*}{\centering\rotatebox[origin=c]{60}{unmasking}}} \\
    \multicolumn{7}{c}{} \\
    \multicolumn{7}{c}{} \\
    \multicolumn{7}{c}{} \\
    \multicolumn{7}{c}{} \\

    \midrule
    \multicolumn{2}{l}{} & [\textit{F1 score}] & [\textit{binary-accuracy}] & [\textit{RMSLE}] & \multicolumn{2}{c}{[\textit{top-$5$-RMSE}]} \\
    \midrule
        \multirow{2}{*}{MLP}                            & oeis   & 0.33 & 0.733 & 0.597 & 0.301 & 2.918 \\
                                                        & synth  & 0.43 & 0.943 & 0.430 & 1.690 & 3.408 \\
        \cmidrule(r){3-7}           
        \multirow{2}{*}{RNN}                            & oeis   & \besto{0.37} & \besto{0.869} & 0.603 & 0.383 & 2.944 \\
                                                        & synth  & \bests{0.53} & \bests{0.984} & 0.406 & 0.438 & 3.379 \\
        \cmidrule(r){3-7}           
        \multirow{2}{*}{CNN}                            & oeis   & 0.22 & 0.551 & 0.733 & 0.428 & \besto{2.440} \\
                                                        & synth  & 0.39 & 0.900 & 0.579 & 0.643 & \bests{2.812} \\
        \cmidrule(r){3-7}           
        \multirow{2}{*}{Transformer}                    & oeis   & 0.33 & 0.736 & \besto{0.578} & \besto{0.267} & 2.811 \\
                                                        & synth  & 0.44 & 0.938 & \bests{0.395} & \bests{0.270} & 3.091 \\
        \cmidrule(r){3-7}           
        \multirow{2}{*}{k-Nearest Neighbours}           & oeis   & 0.33 &   --  & 0.808 &   --  &   --  \\
                                                        & synth  & 0.41 &   --  & 0.486 &   --  &   --  \\
        \cmidrule(r){3-7}           
        \multirow{2}{*}{Gaussian Naive Bayes}           & oeis   & 0.23 &   --  &   --  &   --  &   --  \\
                                                        & synth  & 0.37 &   --  &   --  &   --  &   --  \\
        \cmidrule(r){3-7}           
        \multirow{2}{*}{Support Vector Machine}         & oeis   & 0.31 &   --  &   --  &   --  &   --  \\
                                                        & synth  & 0.35 &   --  &   --  &   --  &   --  \\
        \cmidrule(r){3-7}           
        \multirow{2}{*}{Decision Tree}                  & oeis   & 0.36 &   --  & 0.730 &   --  &   --  \\
                                                        & synth  & 0.49 &   --  & 0.427 &   --  &   --  \\
        \cmidrule(r){3-7}           
        \multirow{2}{*}{Random Forest}                  & oeis   & 0.34 &   --  & 0.730 &   --  &   --  \\
                                                        & synth  & 0.51 &   --  & 0.427 &   --  &   --  \\
        \cmidrule(r){3-7}   
        \multirow{2}{*}{Grad.-Boosted Rand. Forest}     & oeis   & 0.27 &   --  & 0.702 &   --  &   --  \\
                                                        & synth  & 0.40 &   --  & 0.484 &   --  &   --  \\
        \cmidrule(r){3-7}           
        \multirow{2}{*}{AdaBoost}                       & oeis   & 0.31 &   --  & 0.842 &   --  &   --  \\
                                                        & synth  & 0.38 &   --  & 0.662 &   --  &   --  \\
        \cmidrule(r){3-7}           
        \multirow{2}{*}{XGBoost}                        & oeis   & 0.37 &   --  & 0.719 &   --  &   --  \\
                                                        & synth  & 0.51 &   --  & 0.433 &   --  &   --  \\
        \cmidrule(r){3-7}           
        \multirow{2}{*}{Elastic Net Regressor}          & oeis   &  --  &   --  & 0.814 &   --  &   --  \\
                                                        & synth  &  --  &   --  & 0.716 &   --  &   --  \\
        \cmidrule(r){3-7}           
        \multirow{2}{*}{Ridge Regressor}                & oeis   &  --  &   --  & 0.797 &   --  &   --  \\
                                                        & synth  &  --  &  --   & 0.682 &   --  &   --  \\
        \cmidrule(r){3-7}           
        \multirow{2}{*}{Lasso Regressor}                & oeis   &  --  &   --  & 0.827 &   --  &   --  \\
                                                        & synth  &  --  &   --  & 0.747 &   --  &   --  \\
        \cmidrule(r){3-7}           
        \multirow{2}{*}{Linear Regressor}               & oeis   &  --  &   --  & 0.797 &   --  &   --  \\
                                                        & synth  &  --  &   --  & 0.682 &   --  &   --  \\
        \cmidrule(r){3-7}           
        \multirow{2}{*}{Dummy}                          & oeis   & 0.50 &   --  & 0.923 &   --  &   --  \\
                                                        & synth  & 0.50 &   --  & 0.877 &   --  &   --  \\
    \bottomrule 
    
    \end{tabular}
}
\caption{
    An overview of the results for all tasks, evaluated across the whole dataset.
    MLP, RNN, and CNN stand for multi-layer perceptron, recurrent neural network, and convolutional neural network.
    \besto{Emphasis} and \bests{emphasis} mark the best performing models for the OEIS and synthetic data, respectively.
    For F1 score and binary accuracy, higher is better. For RMSLE and top-5-RMSE, lower is better.
}
\label{table:main-results}
\end{table*}

\subsection{Metric Interpretation}
The macro-averaged F1-scores for the \textit{sequence classification} task in \Cref{table:main-results} suggest that even the best-performing models have an average score of little over $0.5$. The F1 score of $0.5$ can be achieved in many ways, but would for example correspond to a precision-recall performance of $0.5-0.5$.
In the \textit{next sequence-part prediction}, the RNNs appear to be nearing the perfect accuracy score of $1.0$, though some room for improvement remains to be seen in the case of the organic dataset.
The root mean squared logarithmic errors of $0.578$ (best OEIS performance) and $0.395$ (best synthetic performance) that appear in the results for the \textit{continuation} task correspond to uniform difference of logarithms of the same magnitude.
In contrast, the two respective worst performances among the baselines models correspond to uniform difference between logarithms of $0.923,0.877$.

The top-$5$ root mean squared errors are more straightforward to interpret.
One would arrive at RMSE between initial sequence segments $s_1,s_2$ if $s_1$ were larger or smaller than $s_2$ by exactly $5$ in every one of its elements.
Top-$5$-RMSEs of $0.267,0.270$ for the \textit{sequence similarity} task therefore correspond to mean uniform difference of the same magnitude in every element when comparing the true sequence to the best fit from among the top $5$ results of the similarity search.
The same metric is used for the \textit{unmasking} task.
In the light of the distribution of the elements of the sequences considered covering values from $0$ to several million, the performances of baseline models, especially in the similarity task, appear to be remarkably strong.

\subsection{Comparative Analysis}
Reviewing the results of \Cref{table:main-results} and further \Cref{appendix:baseline-performances}, we note that even just the performance of baseline models on the synthetic dataset is often quite strong in absolute terms.
We also notice a general trend among all models to perform better on synthetic data than on the organic OEIS sets.
This is not unexpected, since the OEIS data is highly varied and comes from a large variety of sources, whereas the synthetic data is generated according to a strict, uniform procedure, thus having a more regular distribution.
Nevertheless, we observe that training on synthetic data alone still yields solid performance on the organic dataset across all models.

In the \textit{classification} task, RNNs and random forests achieve the best results across all categories.
Unsurprisingly, RNNs very accurately identify bounded and increasing sequences, while random forests lead for modulo, prime, exponential and trigonometric sets.
RNNs also show a consistent lead for the \textit{next sequence-part prediction}.
Transformers and CNNs dominate the results for \textit{similarity} under the top-$k$ accuracy, with the exception of organic periodic functions, which are best handled by recurrent networks.
Transformers alone lead in the same task when evaluating by top-$k$ RMSLE, and likewise for convolutional networks in the \textit{unmasking}task .

The best performances for the \textit{continuation} task are almost evenly split between recurrent networks and transformers, where RNNs lead for polynomial, exponential, trigonometric, and periodic sequences (transformers being the worst performers).
Transformers yield the best results modulo, prime, bounded, increasing, and all sequences together.

We lay out our expectations for model performance on this benchmark, also in relation to human ability, in \Cref{appendix:expected-machine-performance}.

\section{Related Work}
\label{section:relatedWork}
There has recently been a noticeable movement in neural network research towards understanding how DNNs learn to abstract.
On the side of investigation, traditional architectures were analysed by \cite{kozma2018evolution} in terms of the emergence of knowledge across the network, a well-defined metric for the generalisation ability of neural networks was introduced in \cite{gain2020abstraction}, and a methodology to assess knowledge representation in deep neural networks trained for object recognition in computer vision was proposed in \cite{ilin2017abstraction}.
The increasing interest in the learning of abstractions also prompted the incorporation of relevant inductive biases into deep neural architectures and training curricula, as was seen in concept acquisition \cite{wu2016value, higgins2017scan}, with the introduction of the neural state machine \cite{hudson2019learning} in computer vision, and causal abstraction analysis \cite{beckers2019abstracting, beckers2020approximate, geiger2021causal} in natural language inference.
Further, efforts have already begun to assess the capability of models to perform abstracting visual reasoning \cite{zhang2019raven,barrett2018measuring,zhang2020machine,chollet2019measure}.

A common, classical, and still challenging instance of abstraction learning in the context of number sequences is the task of \textit{symbolic regression}.
A number of genetic programming models and purpose-specific datasets have been proposed in the field in its over 30 years of existence, and a systematising benchmark, SRBench \cite{la2021contemporary}, was recently introduced.
It combines 130 smaller numerical datasets, both organically grown and synthetically generated, with the PMLB \cite{olson2017pmlb}, and comes with an evaluation of a range of symbolic regression models using a newly-proposed metric.
The inherent focus of the task on interpretability makes it suitable for industrial use but leads to challenges in identification of prevailing generative concepts, as two sequences originating from two distinct instances of the same rule may be best fitted by two formulas completely different in their nature.

Focusing on \textit{integer sequences}, the Online Encyclopedia of Integer Sequences (OEIS) was presented in \cite{sloane2007line}.
The entries of the encyclopedia come from both individual human contributors and automated mechanisms for the invention of ``interesting'' sequences \cite{colton2000automatic}.
It was used for sequence classification combining heuristics and machine learning methods in \cite{wu2018can}, for the sequence continuation task by fully-connected neural networks in \cite{ragni2011predicting}, for digit-level sequence term regression to highlight the computational limits of neural networks in \cite{nam2019number}, and for the learning of mathematical properties of integers for use in natural language processing by training OEIS-sequence embeddings in \cite{ryskina2021learning}.
The latest version, OEISv4, is the most comprehensive source of annotated information on integer sequences, containing over \num{300000} entries.
The dataset has further seen use in the emergent sub-area of deep symbolic regression \cite{lample2019deep, petersen2019deep, d2022deep, kamienny2022end}.

Our experience shows that the OEIS data is too sparse and too closely tailored to the needs of human reader to be useful for training of machine learning models for integer sequence comprehension. It can still, however, serve as an interesting proxy of ``usefulness'' (such as in \cite{colton2000automatic}) in model evaluation when appropriately filtered and pre-processed for that purpose.

\section{Avenues for Future Work}
\label{section:discussion}
The carrying advantage of the focus on integer sequences instead of other -- potentially richer -- input modalities is that we can directly interpret the performance scores of individual models as their ability to comprehend sequence-giving abstractions, and have the confidence that no model performance has been hampered by its insufficient understanding of the input representation.
Here, while the models we evaluated appear to show some level of understanding of patterns underlying integer sequences, there is still significant room for improvement, especially in multi-class classification, sequence continuation across all classes, and sequence unmasking.

The results of \Cref{section:baselines} were all produced for a ``static'' mode of operation, in which all of the query data (e.g. the sequence to classify or a sequence prefix to continue) was provided to the model upfront and as a whole.
A mode of operation occurring perhaps more naturally in most practical scenarios is that in which the model is active in its learning and interacts with an oracle, polling for information until it is confident that it can provide and answer.
Such interactive variants can be readily formulated for all tasks in \Cref{section:task-types}, but require a more sophisticated set of evaluation metrics that takes into consideration the amount of information the model requested before producing an answer.
We believe that this setup deserves attention as it can provide valuable insights into model reasoning, and we aim to tackle it in our future work.

\section{Conclusion}
\label{section:conclusion}
Integer sequences frequently arise as the natural representation form for finitary phenomena.
Focusing on integer sequences allows us to directly address the problem of learning abstractions and removes the otherwise necessary overhead of learning modality-specific representations.

The benchmarking toolkit for integer sequences presented in this work is by design general in its purpose, aimed at fundamental understanding due to its use of integers as primitive representations, and hierarchically encompasses many of the tasks that have previously appeared isolated in the literature.

It is our hope that our work will help attract attention to the challenges of designing models that perceive logical relationships ruling over the training corpora and reason during inference, thus helping to facilitate future advancements on the frontiers of general artificial intelligence.

\bibliographystyle{plainnat}
\bibliography{bibliography}

\begin{thebibliography}{48}
\providecommand{\natexlab}[1]{#1}
\providecommand{\url}[1]{\texttt{#1}}
\expandafter\ifx\csname urlstyle\endcsname\relax
  \providecommand{\doi}[1]{doi: #1}\else
  \providecommand{\doi}{doi: \begingroup \urlstyle{rm}\Url}\fi

\bibitem[Barrett et~al.(2018)Barrett, Hill, Santoro, Morcos, and
  Lillicrap]{barrett2018measuring}
David Barrett, Felix Hill, Adam Santoro, Ari Morcos, and Timothy Lillicrap.
\newblock Measuring abstract reasoning in neural networks.
\newblock In \emph{International conference on machine learning}, pages
  511--520. PMLR, 2018.

\bibitem[Beckers and Halpern(2019)]{beckers2019abstracting}
Sander Beckers and Joseph~Y Halpern.
\newblock Abstracting causal models.
\newblock In \emph{Proceedings of the aaai conference on artificial
  intelligence}, volume~33, pages 2678--2685, 2019.

\bibitem[Beckers et~al.(2020)Beckers, Eberhardt, and
  Halpern]{beckers2020approximate}
Sander Beckers, Frederick Eberhardt, and Joseph~Y Halpern.
\newblock Approximate causal abstractions.
\newblock In \emph{Uncertainty in Artificial Intelligence}, pages 606--615.
  PMLR, 2020.

\bibitem[{Belcak, Peter} et~al.(2022{\natexlab{a}}){Belcak, Peter}, {Kastrati,
  Ard}, and {Schenker, Flavio}]{fact_benchmarking_baselines_through_doi}
{Belcak, Peter}, {Kastrati, Ard}, and {Schenker, Flavio}.
\newblock Fact benchmarking - the benchmarking baseline models of the finitary
  abstraction comprehension toolkit, 2022{\natexlab{a}}.
\newblock URL \url{https://doi.org/10.3929/ethz-b-000565644}.

\bibitem[{Belcak, Peter} et~al.(2022{\natexlab{b}}){Belcak, Peter}, {Kastrati,
  Ard}, and {Schenker, Flavio}]{fact_dataset_through_doi}
{Belcak, Peter}, {Kastrati, Ard}, and {Schenker, Flavio}.
\newblock Fact dataset - the dataset of the finitary abstraction comprehension
  toolkit, 2022{\natexlab{b}}.
\newblock URL \url{https://doi.org/10.3929/ETHZ-B-000562705}.

\bibitem[{Belcak, Peter} et~al.(2022{\natexlab{c}}){Belcak, Peter}, {Kastrati,
  Ard}, and {Schenker, Flavio}]{factlib_through_doi}
{Belcak, Peter}, {Kastrati, Ard}, and {Schenker, Flavio}.
\newblock Factlib - the library of the finitary abstraction comprehension
  toolkit, 2022{\natexlab{c}}.
\newblock URL \url{https://doi.org/10.3929/ethz-b-000565638}.

\bibitem[Chollet(2019)]{chollet2019measure}
Fran{\c{c}}ois Chollet.
\newblock On the measure of intelligence.
\newblock \emph{arXiv preprint arXiv:1911.01547}, 2019.

\bibitem[Colton et~al.(2000)Colton, Bundy, and Walsh]{colton2000automatic}
Simon Colton, Alan Bundy, and Toby Walsh.
\newblock Automatic invention of integer sequences.
\newblock In \emph{AAAI/IAAI}, pages 558--563, 2000.

\bibitem[d'Ascoli et~al.(2022)d'Ascoli, Kamienny, Lample, and
  Charton]{d2022deep}
St{\'e}phane d'Ascoli, Pierre-Alexandre Kamienny, Guillaume Lample, and
  Fran{\c{c}}ois Charton.
\newblock Deep symbolic regression for recurrent sequences.
\newblock \emph{arXiv preprint arXiv:2201.04600}, 2022.

\bibitem[Devlin et~al.(2018)Devlin, Chang, Lee, and Toutanova]{devlin2018bert}
Jacob Devlin, Ming-Wei Chang, Kenton Lee, and Kristina Toutanova.
\newblock Bert: Pre-training of deep bidirectional transformers for language
  understanding.
\newblock \emph{arXiv preprint arXiv:1810.04805}, 2018.

\bibitem[Gain and Siegelmann(2020)]{gain2020abstraction}
Alex Gain and Hava Siegelmann.
\newblock Abstraction mechanisms predict generalization in deep neural
  networks.
\newblock In \emph{International Conference on Machine Learning}, pages
  3357--3366. PMLR, 2020.

\bibitem[Gebru et~al.(2018)Gebru, Morgenstern, Vecchione, Vaughan, Wallach,
  Daum{\'e}~III, and Crawford]{gebru2018datasheets}
Timnit Gebru, Jamie Morgenstern, Briana Vecchione, Jennifer~Wortman Vaughan,
  Hanna Wallach, Hal Daum{\'e}~III, and Kate Crawford.
\newblock Datasheets for datasets.
\newblock \emph{arXiv preprint arXiv:1803.09010}, 2018.

\bibitem[Geiger et~al.(2021)Geiger, Lu, Icard, and Potts]{geiger2021causal}
Atticus Geiger, Hanson Lu, Thomas Icard, and Christopher Potts.
\newblock Causal abstractions of neural networks.
\newblock \emph{Advances in Neural Information Processing Systems}, 34, 2021.

\bibitem[Higgins et~al.(2017)Higgins, Sonnerat, Matthey, Pal, Burgess, Bosnjak,
  Shanahan, Botvinick, Hassabis, and Lerchner]{higgins2017scan}
Irina Higgins, Nicolas Sonnerat, Loic Matthey, Arka Pal, Christopher~P Burgess,
  Matko Bosnjak, Murray Shanahan, Matthew Botvinick, Demis Hassabis, and
  Alexander Lerchner.
\newblock Scan: Learning hierarchical compositional visual concepts.
\newblock \emph{arXiv preprint arXiv:1707.03389}, 2017.

\bibitem[Hochreiter and Schmidhuber(1997)]{lstm}
Sepp Hochreiter and Jürgen Schmidhuber.
\newblock Long short-term memory.
\newblock \emph{Neural computation}, 9:\penalty0 1735--80, 12 1997.
\newblock \doi{10.1162/neco.1997.9.8.1735}.

\bibitem[Hudson and Manning(2019)]{hudson2019learning}
Drew Hudson and Christopher~D Manning.
\newblock Learning by abstraction: The neural state machine.
\newblock \emph{Advances in Neural Information Processing Systems}, 32, 2019.

\bibitem[Husain et~al.(2019)Husain, Wu, Gazit, Allamanis, and
  Brockschmidt]{husain2019codesearchnet}
Hamel Husain, Ho-Hsiang Wu, Tiferet Gazit, Miltiadis Allamanis, and Marc
  Brockschmidt.
\newblock Codesearchnet challenge: Evaluating the state of semantic code
  search.
\newblock \emph{arXiv preprint arXiv:1909.09436}, 2019.

\bibitem[Ilin et~al.(2017)Ilin, Watson, and Kozma]{ilin2017abstraction}
Roman Ilin, Thomas Watson, and Robert Kozma.
\newblock Abstraction hierarchy in deep learning neural networks.
\newblock In \emph{2017 International Joint Conference on Neural Networks
  (IJCNN)}, pages 768--774. IEEE, 2017.

\bibitem[Kamienny et~al.(2022)Kamienny, d'Ascoli, Lample, and
  Charton]{kamienny2022end}
Pierre-Alexandre Kamienny, St{\'e}phane d'Ascoli, Guillaume Lample, and
  Fran{\c{c}}ois Charton.
\newblock End-to-end symbolic regression with transformers.
\newblock \emph{arXiv preprint arXiv:2204.10532}, 2022.

\bibitem[Kolmogorov(1965)]{kolmogorov}
Andrei~N Kolmogorov.
\newblock Three approaches to the quantitative definition of information'.
\newblock \emph{Problems of information transmission}, 1\penalty0 (1):\penalty0
  1--7, 1965.

\bibitem[Kozma et~al.(2018)Kozma, Ilin, and Siegelmann]{kozma2018evolution}
Robert Kozma, Roman Ilin, and Hava~T Siegelmann.
\newblock Evolution of abstraction across layers in deep learning neural
  networks.
\newblock \emph{Procedia computer science}, 144:\penalty0 203--213, 2018.

\bibitem[La~Cava et~al.(2021)La~Cava, Orzechowski, Burlacu, de~Fran{\c{c}}a,
  Virgolin, Jin, Kommenda, and Moore]{la2021contemporary}
William La~Cava, Patryk Orzechowski, Bogdan Burlacu, Fabr{\'\i}cio~Olivetti
  de~Fran{\c{c}}a, Marco Virgolin, Ying Jin, Michael Kommenda, and Jason~H
  Moore.
\newblock Contemporary symbolic regression methods and their relative
  performance.
\newblock \emph{arXiv preprint arXiv:2107.14351}, 2021.

\bibitem[Lample and Charton(2019)]{lample2019deep}
Guillaume Lample and Fran{\c{c}}ois Charton.
\newblock Deep learning for symbolic mathematics.
\newblock \emph{arXiv preprint arXiv:1912.01412}, 2019.

\bibitem[Mikolov et~al.(2013)Mikolov, Chen, Corrado, and
  Dean]{mikolov2013efficient}
Tomas Mikolov, Kai Chen, Greg Corrado, and Jeffrey Dean.
\newblock Efficient estimation of word representations in vector space.
\newblock \emph{arXiv preprint arXiv:1301.3781}, 2013.

\bibitem[Nam et~al.(2019)Nam, Kim, and Jung]{nam2019number}
Hyoungwook Nam, Segwang Kim, and Kyomin Jung.
\newblock Number sequence prediction problems for evaluating computational
  powers of neural networks.
\newblock In \emph{Proceedings of the AAAI Conference on Artificial
  Intelligence}, volume~33, pages 4626--4633, 2019.

\bibitem[Olson et~al.(2017)Olson, La~Cava, Orzechowski, Urbanowicz, and
  Moore]{olson2017pmlb}
Randal~S Olson, William La~Cava, Patryk Orzechowski, Ryan~J Urbanowicz, and
  Jason~H Moore.
\newblock Pmlb: a large benchmark suite for machine learning evaluation and
  comparison.
\newblock \emph{BioData mining}, 10\penalty0 (1):\penalty0 1--13, 2017.

\bibitem[Pedregosa et~al.(2011)Pedregosa, Varoquaux, Gramfort, Michel, Thirion,
  Grisel, Blondel, Prettenhofer, Weiss, Dubourg, Vanderplas, Passos,
  Cournapeau, Brucher, Perrot, and Duchesnay]{scikit}
F.~Pedregosa, G.~Varoquaux, A.~Gramfort, V.~Michel, B.~Thirion, O.~Grisel,
  M.~Blondel, P.~Prettenhofer, R.~Weiss, V.~Dubourg, J.~Vanderplas, A.~Passos,
  D.~Cournapeau, M.~Brucher, M.~Perrot, and E.~Duchesnay.
\newblock Scikit-learn: Machine learning in {P}ython.
\newblock \emph{Journal of Machine Learning Research}, 12:\penalty0 2825--2830,
  2011.

\bibitem[Penrose(1990)]{penrose1990emperor}
Roger Penrose.
\newblock \emph{The Emperor's New Mind: Concerning Computers, Minds, and the
  Laws of Physics}.
\newblock Viking Penguin, 1990.
\newblock ISBN 0140145346.

\bibitem[Petersen et~al.(2019)Petersen, Larma, Mundhenk, Santiago, Kim, and
  Kim]{petersen2019deep}
Brenden~K Petersen, Mikel~Landajuela Larma, T~Nathan Mundhenk, Claudio~P
  Santiago, Soo~K Kim, and Joanne~T Kim.
\newblock Deep symbolic regression: Recovering mathematical expressions from
  data via risk-seeking policy gradients.
\newblock \emph{arXiv preprint arXiv:1912.04871}, 2019.

\bibitem[Ragni and Klein(2011)]{ragni2011predicting}
Marco Ragni and Andreas Klein.
\newblock Predicting numbers: an ai approach to solving number series.
\newblock In \emph{Annual Conference on Artificial Intelligence}, pages
  255--259. Springer, 2011.

\bibitem[Russell and Carter(2003)]{russell2003times}
Kenneth~A Russell and Philip~J Carter.
\newblock \emph{The Times book of IQ tests}, volume~3.
\newblock Kogan Page Publishers, 2003.

\bibitem[Ryskina and Knight(2021)]{ryskina2021learning}
Maria Ryskina and Kevin Knight.
\newblock Learning mathematical properties of integers.
\newblock \emph{arXiv preprint arXiv:2109.07230}, 2021.

\bibitem[Schmidhuber(1997)]{schmidhuber}
Jürgen Schmidhuber.
\newblock Discovering neural nets with low kolmogorov complexity and high
  generalization capability.
\newblock \emph{Neural Networks}, 10\penalty0 (5):\penalty0 857--873, 1997.
\newblock ISSN 0893-6080.
\newblock \doi{https://doi.org/10.1016/S0893-6080(96)00127-X}.
\newblock URL
  \url{https://www.sciencedirect.com/science/article/pii/S089360809600127X}.

\bibitem[Sigler(2003)]{sigler2003fibonacci}
Laurence Sigler.
\newblock \emph{Fibonacci’s Liber Abaci: a translation into modern English of
  Leonardo Pisano’s book of calculation}.
\newblock Springer Science \& Business Media, 2003.

\bibitem[Sloane({\natexlab{a}})]{oeis}
Neil J.~A. Sloane.
\newblock The on-line encyclopedia of integer sequences, {\natexlab{a}}.
\newblock URL \url{https://oeis.org/}.

\bibitem[Sloane({\natexlab{b}})]{oeis_keywords}
Neil J.~A. Sloane.
\newblock Oeis keywords, {\natexlab{b}}.
\newblock URL \url{https://oeis.org/wiki/Keywords}.

\bibitem[Sloane({\natexlab{c}})]{oeis_style_sheet}
Neil J.~A. Sloane.
\newblock Style sheet for contributers., {\natexlab{c}}.
\newblock URL \url{https://oeis.org/wiki/Style_Sheet}.

\bibitem[Sloane(1973)]{book_handbook}
Neil J.~A. Sloane.
\newblock \emph{A Handbook of Integer Sequences}.
\newblock Academic Press, 1973.
\newblock ISBN 0-12-648550-X.

\bibitem[Sloane(2007)]{sloane2007line}
Neil J.~A. Sloane.
\newblock The on-line encyclopedia of integer sequences.
\newblock In \emph{Towards mechanized mathematical assistants}, pages 130--130.
  Springer, 2007.

\bibitem[Sloane and Plouffe(1995)]{book_encyclopedia}
Neil J.~A. Sloane and S.~Plouffe.
\newblock \emph{The Encyclopedia of Integer Sequences}.
\newblock Academic Press, 1995.
\newblock ISBN 0-12-558630-2.

\bibitem[Solomonoff(1986)]{solomonoff}
Ray Solomonoff.
\newblock The application of algorithmic probability to problems in artificial
  intelligence.
\newblock In Laveen~N. KANAL and John~F. LEMMER, editors, \emph{Uncertainty in
  Artificial Intelligence}, volume~4 of \emph{Machine Intelligence and Pattern
  Recognition}, pages 473--491. North-Holland, 1986.
\newblock \doi{https://doi.org/10.1016/B978-0-444-70058-2.50040-1}.
\newblock URL
  \url{https://www.sciencedirect.com/science/article/pii/B9780444700582500401}.

\bibitem[Stranneg{\aa}rd et~al.(2013)Stranneg{\aa}rd, Amirghasemi, and
  Ulfsb{\"a}cker]{strannegaard2013anthropomorphic}
Claes Stranneg{\aa}rd, Mehrdad Amirghasemi, and Simon Ulfsb{\"a}cker.
\newblock An anthropomorphic method for number sequence problems.
\newblock \emph{Cognitive Systems Research}, 22:\penalty0 27--34, 2013.

\bibitem[Team()]{sympy}
SymPy~Development Team.
\newblock Sympy.
\newblock URL \url{https://www.sympy.org/en/index.html}.

\bibitem[Vaswani et~al.(2017)Vaswani, Shazeer, Parmar, Uszkoreit, Jones, Gomez,
  Kaiser, and Polosukhin]{transformer}
Ashish Vaswani, Noam Shazeer, Niki Parmar, Jakob Uszkoreit, Llion Jones,
  Aidan~N. Gomez, Lukasz Kaiser, and Illia Polosukhin.
\newblock Attention is all you need.
\newblock \emph{CoRR}, abs/1706.03762, 2017.
\newblock URL \url{http://arxiv.org/abs/1706.03762}.

\bibitem[Wu(2018)]{wu2018can}
Chai~Wah Wu.
\newblock Can machine learning identify interesting mathematics? an exploration
  using empirically observed laws.
\newblock \emph{arXiv preprint arXiv:1805.07431}, 2018.

\bibitem[Wu et~al.(2016)Wu, Shen, Liu, Dick, and Van Den~Hengel]{wu2016value}
Qi~Wu, Chunhua Shen, Lingqiao Liu, Anthony Dick, and Anton Van Den~Hengel.
\newblock What value do explicit high level concepts have in vision to language
  problems?
\newblock In \emph{Proceedings of the IEEE conference on computer vision and
  pattern recognition}, pages 203--212, 2016.

\bibitem[Zhang et~al.(2019)Zhang, Gao, Jia, Zhu, and Zhu]{zhang2019raven}
Chi Zhang, Feng Gao, Baoxiong Jia, Yixin Zhu, and Song-Chun Zhu.
\newblock Raven: A dataset for relational and analogical visual reasoning.
\newblock In \emph{Proceedings of the IEEE/CVF Conference on Computer Vision
  and Pattern Recognition}, pages 5317--5327, 2019.

\bibitem[Zhang et~al.(2020)Zhang, Zhang, Zhu, and Zhu]{zhang2020machine}
Wenhe Zhang, Chi Zhang, Yixin Zhu, and Song-Chun Zhu.
\newblock Machine number sense: A dataset of visual arithmetic problems for
  abstract and relational reasoning.
\newblock In \emph{Proceedings of the AAAI Conference on Artificial
  Intelligence}, volume~34, pages 1332--1340, 2020.

\end{thebibliography}

\clearpage
\appendix

\section*{Acknowledgments}
We thank Emanuel Jampen. His work on initial processing of the OEIS dataset showed the feasibility of this project. We also thank Neil Sloane for founding the OEIS project and sharing it in public domain.

\newpage

\section{OEIS Processing}
\label{appendix:oeis}
The On-Line Encyclopedia of Integer Sequences\cite{oeis}, OEIS for short, is an online dataset of integer sequences. Founded in $1996$ by Neil Sloane the dataset was since then steadily expanded by amateur and professional mathematicians as well. In $18$ columns the dataset provides information about various properties of those sequences. The 18 column names are briefly explained in \Cref{subsection:dataset_fields}. We worked with an offline version of the OEIS dataset which contains $342\,304$ sequences. On a content level they range from well known like $A000045$, Fibonacci numbers: $F(n) = F(n-1) + F(n-2)$  with $F(0) = 0$ and $F(1) = 1$, over simple ones like, $A000004$: the zero sequence and complex ones like $A339123$: number of 2-connected multigraphs with $n$ edges and rooted at two indistinguishable vertices and have no decomposition into parallel components rooted at the two distinguished vertices, to quite creative ones like $A001049$: numbered stops in Manhattan on the Lexington Avenue subway. A detailed summary can be found in \cite{sloane2007line}.

\paragraph{Field Names}\label{subsection:dataset_fields}

In this section we give a brief explanation of the 18 fields in the OEIS dataset. This is based on the style sheet provided by OEIS \cite{oeis_style_sheet}.

\begin{itemize}
  \item \textbf{oeis\_id}
  
    A unique 6 digit number preceded by an "A". For example \textit{A005735}
  
  \item \textbf{identification}
  
    This refers to the ID the given sequence had in one of the books \textit{A Handbook of Integer Sequences}\cite{book_handbook} (M followed by a 4-digit number) or \textit{The Encyclopedia of Integer Sequences}\cite{book_encyclopedia} (N followed by a 4-digit number).

  \item \textbf{value\_list}
  
  A list of comma separated integers. The actual sequence of interest. Depending on the sequence in question the length of this list can vary by some orders of magnitude. For example $A058445$ contains only one element while the value list of $A175320$ has a length of $1\,578\,727$.
  
  \item \textbf{name}
  
  A brief explanation of the sequence. Sometimes, when possible, this already contains an easy-to-use formula to generate the sequence. For example $A005843$ has the name \textit{The nonnegative even numbers: a(n) = 2n.}
  
  \item \textbf{comments}
  
  Further general details and side-notes about the sequence that would make the name too long. Here we often can find alternative formulas to generate the sequence or different places in all of mathematics where this sequence pops up. For example one of the comments of sequence $A000045$, the Fibonacci Numbers, is \textit{"Also the number of independent vertex sets and vertex covers in the (n-2)-path graph."}
  
  \item \textbf{detailed\_references}
  
  References to journal papers and books that can not be linked in the "links" field.
  \item \textbf{links}
  
  References to material which can be accessed online.
  \item \textbf{formulas}
  
  Generating functions, closed formulas and other methods to calculate the sequence. 
  \item \textbf{examples}
  
  Examples of how to find a term of the sequence and how to interpret its value.
  \item \textbf{maple\_programs}
  
  Programs written in maple to generate elements of the sequence.
  \item \textbf{mathematica\_programs}
  
  Programs written in mathematica to generate elements of the sequence.
  \item \textbf{other\_programs}
  
  Programs written in programming languages other than maple or mathematica (for example python) to generate elements of the sequence.
  \item \textbf{cross\_reference}
  
  References to other sequences in the dataset which are related in some way.
  \item \textbf{keywords}
  
  Keywords from a short set of possibilities. For example \textit{nonn} is used for sequences that have no negative values currently in their respective \textit{value\_list} field. 
  
  \item \textbf{offset\_a}
  
  Index of the first element in the \textit{value\_list}. For example $A005843$, the aforementioned sequence of \textit{"The nonnegative even numbers: a(n) = 2n."}, has an offset\_a of $0$ because the first element is calculated by $a(0) = 2*0$, i.e. the index is $0$.
  
  \item \textbf{offset\_b}
  
  Index of the first element that has an absolute value larger than $1$.
  \item \textbf{author}
  
  Name of the original contributor and date of first contribution.
  \item \textbf{extensions\_and\_errors}
  
  Used to claim credit for additions to the entry that can't be properly acknowledged in other fields.
  
\end{itemize}

\subsection{Characteristics}

In this section, we introduce some other characteristics of the dataset that required preprocessing and extensions.
    
\begin{itemize}

    \item \textbf{NULL Values}
    
    Not every sequence contains information in every of the $18$ columns. Table \ref{table:valid_entries} shows the number of valid entries per column in the dataset. We say an entry is valid if it is not \textit{NULL}. When programming classification methods these invalid entries have to be taken into consideration. Of course this can be a limitation for the meaningfulness of such classification methods. For example we can not check whether the \textit{formulas} field contains a Fibonacci-Like formula when it is NULL. However, the fact that the field is NULL can also be a hint that no known closed formula exists (otherwise someone would probably contribute it to the OEIS).

    \begin{table}
        \centering
            \begin{tabular}{| l| l| l|}
                 \hline
                 Field name & Valid entries & Invalid entries\\
                 \hline
                oeis\_id & 342304 & 0\\
                identification & 5533 & 336771\\
                value\_list & 341889 & 415\\
                name & 342304 & 0\\
                comments & 198198 & 144106\\
                detailed\_references & 33687 & 308617\\
                links & 237038 & 105266\\
                formulas & 149606 & 192698\\
                examples & 163666 & 178638\\
                maple\_programs & 53333 & 288971\\
                mathematica\_programs & 168158 & 174146\\
                other\_programs & 121276 & 221028\\
                cross\_references & 245298 & 97006\\
                keywords & 342304 & 0\\
                offset\_a & 341885 & 419\\
                offset\_b & 340578 & 1726\\
                author & 340455 & 1849\\
                extensions\_and\_errors & 77949 & 264355\\
                \hline
            \end{tabular}
        \caption{Number of Valid and Invalid entries per field}
        \label{table:valid_entries}
    \end{table}
    
    \item \textbf{Sequence lengths}
    
    Since it is harder to calculate elements for some sequences than for others, the length of the \textit{value\_list} shows a high variance. Figure \ref{fig:sequence_length_histogram} shows the distribution of sequence lengths over two different scales. As we can see there is a substantial number of sequences with short sequence lengths (say less than 30). Short sequence lengths can pose a problem to classification methods that need more elements for some mathematical calculations. For example we can fit a polynomial of degree $20$ to $21$ datapoints, but then we can not validate the result on additional datapoints.
    
    \begin{figure}[h]
    \centerline{%
        \includegraphics[trim=0 0 0 0,clip,width=0.49\textwidth]{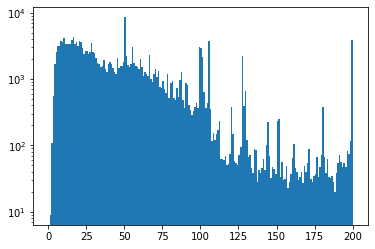}%
        \hspace{20pt}
        \includegraphics[trim=0 0 0 0,clip,width=0.49\textwidth]{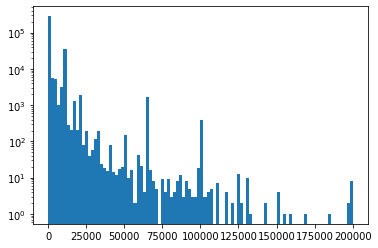}%
    }%
    \vspace{0pt}
    \caption{Distribution of sequence lengths. \textit{Left.} The horizontal axis corresponds to the sequence length, while the vertical axis has logarithmic scale and gives the count of the sequences with such length. \textit{Right.} The same data as on the left but with the roles of axes exchanged. }
    \label{fig:sequence_length_histogram}
\end{figure}

    \item  \textbf{Keyword counts}
    
    Table \ref{table:keyword_counts} shows the number of occurrences of each of the $27$ possible keywords. Again, for a full description of what each keyword means see the corresponding Wiki page from OEIS \cite{oeis_keywords}. 
    
    \begin{table}
        \centering
            \begin{tabular}{| l| l|}
                 \hline
                 Keyword & Occurrences\\
                 \hline
                base & 39097 \\
                bref & 739 \\
                cofr & 2866 \\
                cons & 11454 \\
                core & 178 \\
                dead & 1684 \\
                dumb & 99 \\
                easy & 79148 \\
                eigen & 430 \\
                fini & 6594 \\
                frac & 7050 \\
                full & 5494 \\
                hard & 7137 \\
                hear & 162 \\
                less & 2683 \\
                look & 2829 \\
                more & 20107 \\
                mult & 2120 \\
                nice & 6891 \\
                nonn & 321805 \\
                obsc & 120 \\
                sign & 18480 \\
                tabf & 6442 \\
                tabl & 20248 \\
                unkn & 32 \\
                walk & 4285 \\
                word & 749 \\
                \hline
            \end{tabular}
        \caption{Number of occurrences of each keyword}
        \label{table:keyword_counts}
    \end{table}

\end{itemize}

\subsection{Processing methods}

In order to group similar sequences from OEIS together we  create categories and decide for each sequence whether they belong into this category or not. Since for some sequences it is not immediately clear whether they belong to a given category or not, we relaxed the requirement of this binary classification to a finer graduated classification. To this end, we have $5$ different levels of membership, represented by integers $0$ through $4$. These values indicate how confident our classification is, as seen in the following table:

\begin{table}[h]
\centering
\begin{tabular}{ l l }
 0 & Does likely not belong in this category \\ 
 1 & More likely than not does not belong in this category\\  
 2 & Inconclusive\\
 3 & More likely than not belongs in this category\\
 4 & Does likely belong in this category
\end{tabular}
\end{table}

In this chapter we introduce and explain methods we used to define categories and create the associated labels for membership. Neither our selection of categories nor the different Classification Methods are exhaustive, therefore we implemented an easy to use framework which enables a fast creation of new categories. This framework is explained in the following Section \ref{section:annotator_class}.

\subsection{The Annotator Class}\label{section:annotator_class}
\begin{figure}
    \centering
    \includegraphics[width=0.7\columnwidth]{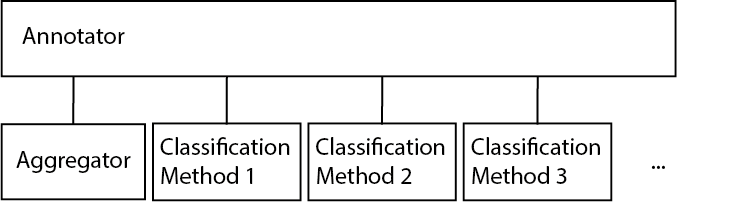}
    \caption{Structure of the Annotator Class with multiple Classification Methods and an Aggregator}
    \label{fig:Annotator}
  \end{figure}

We implemented the Annotator Class which lets a user create new categories of sequences. Each instance of the Annotator Class is responsible for one category. As depicted in Figure \ref{fig:Annotator} each Annotator is linked to an \textit{Aggregator} and one or more \textit{Classification Methods} (see Sections \ref{section:aggregator} and \ref{section:tests} below). The Annotator is responsible for running all its Classification Methods for each sequence, providing the Aggregator with results and storing the 5-Level membership results for future use.

\subsection{Aggregator}\label{section:aggregator}
An Aggregator, as the name suggests, aggregates the results from the Classification Methods and returns a 5-Level membership value.

\subsection{Classification Methods}\label{section:tests}

Classification Methods are functions which take all the information about a sequence (i.e. all the fields described in Section \ref{subsection:dataset_fields}) as an input and return results of some test which are then further processed by the Aggregator. 

Note, all Classification Methods have to check whether the fields which have to be accessed by the test do exist for the given sequence, i.e. the field is not empty or NULL.

As we will see in the following subsections, a Classification Method can be something as simple as searching for a given word in the name of the sequence or more complex like doing some advanced calculations based on the value. However, Classification Methods should be designed to have a rather short runtime per sequence since the same function has to run on all sequences in the dataset. As a rule of thumb, when a test takes one second per sequence the total runtime on the current dataset will be about 4 days.


\paragraph{Mathematical} We used Mathematical Tests to check for properties of the values of the sequences. These methods bear some limitations one needs to be aware off. Some tests require a minimal number of values to return meaningful results. For example a sequence with only $5$ given values is not suitable for fitting a polynomial of degree $10$ to it. On the other hand if the sequence truly would have a polynomial form then elements would be easy to calculate and there would probably be more values provided. Following are descriptions of the mathematical Classification Methods we used.

\begin{itemize}
    \item Newton's Divided Difference
    We used Newton's Divided Difference Method to calculate the degree a polynomial would have to have in order to fit to the values of the sequence. Intuitively this is the discrete signal analog of repeatedly taking the derivative of a polynomial until we end up with the constant $0$ function. 
    By counting the number of iterations, we get the degree a fitting polynomial would need to have, which we can then use for Polynomial Interpolation.
    
    \item Polynomial Interpolation
    
    We used the SymPy\cite{sympy} library to perform symbolic polynomial interpolation with the degree calculated by Newton's Divided Difference Method. This allows us to fit a polynomial of degree $n$ to the first $n+1$ values of the sequence and use the remaining values to calculate the error between the fitted polynomial and the actual values. Of course we need to assert that there are enough values, which were not used during interpolation, to be able to calculate a meaningful error. In cases where the error between the fitted polynomial and the actual values is $0$ and we have enough values, we conclude that the sequence in question indeed has a polynomial form.
    
    \item Exponentials
    
    To detect exponential sequences, such as $A000244$, \textit{Powers of 3}, or $A007689$, \textit{$a(n) = 2^n + 3^n$}, directly from their values $a_i$ we first calculate the sequence of quotients $q$. Each element of $q$ is the quotient of two subsequent elements of the original sequence:
    
    \begin{equation}
        q_i = a_i / a_{i+1}
    \end{equation}
    
    Purely exponential sequences, like \textit{Powers of 3}, produce a constant-valued $q$. In our example $q_i = 1/3$ for all $i$. Such sequences pass this test immediately.
    
    Sequences that are dominated by an exponential like $A006127$, \textit{$a(n) = 2^n + n$}, produce a sequence of quotient that approaches a constant value. In this example we get
    
    \begin{equation}
        \lim_{i\to\infty} q_i = \lim_{i\to\infty}\frac{2^i + i}{2^{i+1} + i+1} = \frac{1}{2}
    \end{equation}
    
    We approximate the limit of $q$ by checking the very last values of it that can be calculated from the given data. If the last $30$ elements are all within some threshold range of the final value we conclude that the sequence probably is dominated by an exponential function. 
    
    \item Primality
    
    Using SymPy again, we tested which sequences can be interpreted as the application of the above classes of functions to prime numbers. The way we implemented this test was limited to numbers smaller than $2^{64}$ (or about $10^{19}$).
    
    \item Boundedness
    
    Testing the absolute values of the sequence elements for boundedness by different values, is an easy but insightful series of tests. 
    
    \item Palindrome

    This test checks whether all elements in a sequence are palindromic, i.e. if all values are numbers that are the same when read from left to right as well as from right to left. Trivially this is true for all sequences that contain only values smaller than $10$ since all single-digit numbers are palindromic (i.e. $7$ is the same number when read from either direction). The results from this test depend on the base of the number system used to display the values of the sequence. For example $12321$ is palindromic in base $10$ but in base $2$ it is not: $12321_{10} = 11000000100001_2$.
    
    \item Periodicity
    
    We tested whether a sequence has a repeating part up to a maximal period length of $10\,000$ elements. To pass the test a sequence needs to have at least $3$ full periods in the \textit{value\_list}.
    
\end{itemize}

\paragraph{String Search} Since most of the fields in the dataset are populated by plain text, one of the easiest methods to get information about a sequence is the search for specific words or strings within a given fields. This method can be a reliable indicator whether a sequence does belong to a category, for example performing a string search of the word \textit{prime} on the name of $A098682$, \textit{Smallest prime larger than $n^n$}, can serve as a good indicator that $A098682$ indeed is at least in some way related to prime numbers. However, the opposite conclusion is harder to draw, is a sequence truly unrelated to prime numbers just because it is not mentioned in the name or comments?

\paragraph{Regular Expressions}Regular Expressions served as a powerful tool to check for patterns in the data. For example we created the regular expression 

\begin{verbatim}
    a\(n\)=[0-9]*\*?a\(n[\+\-][0-9]+\)[\+\-][0-9]*\*?
    a\(n[\+\-][0-9]+\)([\+\-][0-9]*\*?a\(n[\+\-][0-9]+\))*
\end{verbatim}

Which matches character patterns like 
\begin{equation}
    a(n)=2*a(n-3)+5*a(n-5)-17a(n-5)    
\end{equation}

Due to the similarity with the well known Fibonacci Sequence ($A000045$), we call sequences with formulas that are matched by above regular expression \textit{Fibonacci-Like}.



  

\section{Synthetic Generation}
\label{appendix:synthetic}

In this section we explain the steps followed to create synthetic sequences.

From the identified categories (cf. \cref{fig:dataset}) in the OEIS database, we have elected a significant and representative subset and augmented their organic entries with synthetic data.
For each category, a total of 500K synthetic sequences have been generated.
The sequences were generated with logaritmically increasing length of the generating formula.
This way, most of the sequences-generating expressions are relatively short (motivated by Kolmogorov complexity, cf. \Cref{section:dataset}), but longer sequences remain playing a significant role in the dataset.
In the following, we provide information how synthetic extensions of each sequence category are created using context-free grammars. Typically, the ``logical'' terminals in our grammars are $\{Var, Const\}$ and the non-terminals $\{Add, Sub, Mult, NConst, Pow\}$. $NConst$ is defined as the non-terminal which represents the number constants with multiple digits. The reason why we consider it a non-terminal is that we count the length of polynomial based on the number of non-terminals only.

\paragraph{Polynomial} The context-free grammar for polynomials is defined as:
\begin{lstlisting}
T -> Var | Const
N -> Add | Sub | Mult | Pow | NConst | T
Add -> (N + N)
Sub -> (N - N)
Mult -> (N * N)
Pow -> (N ** NConst)
NConst -> (ConstPos NConst) | Const
Var -> x
Const -> 0 | 1 | 2 | 3 | 4 | 5 | 6 | 7 | 8 | 9
ConstPos -> 1 | 2 | 3 | 4 | 5 | 6 | 7 | 8 | 9
\end{lstlisting}

\paragraph{Exponential}
Context-free grammar of exponentials group is defined as:
\begin{lstlisting}
T -> Var | Const
N -> Add | Sub | Mult | Pow | NConst | T
Add -> (N + N)
Sub -> (N - N)
Mult -> (N * N)
Pow -> (N ** N)
NConst -> (ConstPos NConst) | Const
Var -> x
Const -> 0 | 1 | 2 | 3 | 4 | 5 | 6 | 7 | 8 | 9
ConstPos -> 1 | 2 | 3 | 4 | 5 | 6 | 7 | 8 | 9
\end{lstlisting}

\paragraph{Prime}
In this group, we introduce a non terminal $Prime(x)$, which evaluated at value $x \in \{1,2,...\}$ returns the $i^{th}$ prime number. That is, $Prime(1) = 2$, $Prime(2) = 3$, $Prime(3) = 5$, and so on. This way we inject in our sequences prior knowledge about prime numbers. More concretely, the context free grammar is defined as: 

\begin{lstlisting}
T -> Var | Const
N -> Add | Sub | Mult | Pow | NConst | Prime | T
Add -> (N + N)
Sub -> (N - N)
Mult -> (N * N)
Pow -> (N ** N)
NConst -> (ConstPos NConst) | Const
Var -> x
Prime -> prime(x) # defined as a function
Const -> 0 | 1 | 2 | 3 | 4 | 5 | 6 | 7 | 8 | 9
ConstPos -> 1 | 2 | 3 | 4 | 5 | 6 | 7 | 8 | 9
\end{lstlisting}

\paragraph{Periodic}

Periodic group makes use of the previous grammars, but it also has a non terminal that is added in the root of the grammar, which makes the whole sequence repeat only the first few $k$ numbers. More concretely, we can phrase this as a grammar too, by adding primitive $Periodic$ as a root of the grammar 

\begin{lstlisting}
Periodic -> periodic(N, k) # defined as a function
k -> NConst
N -> ... as above ...
\end{lstlisting}
Let sequence $f(x)$ be any sequence, then the function $periodic(f(x), k)$ is defined as $f(x \% k)$.

\paragraph{Modulo}

In this group we add a non terminal modulo and we define the grammar as: 
\begin{lstlisting}
T -> Var | Const
N -> Add | Sub | Mult | Pow | NConst | Modulo | T
Add -> (N + N)
Sub -> (N - N)
Mult -> (N * N)
Pow -> (N ** N)
NConst -> (ConstPos NConst) | Const
Var -> x
Modulo -> N % N
Const -> 0 | 1 | 2 | 3 | 4 | 5 | 6 | 7 | 8 | 9
ConstPos -> 1 | 2 | 3 | 4 | 5 | 6 | 7 | 8 | 9
\end{lstlisting}

\paragraph{Trigonometric}

Here we add two non terminals $Sin$ and $Cos$. The grammar defined in this group can be found in the following.
\begin{lstlisting}
T -> Var | Const
N -> Add | Sub | Mult | Pow | NConst | Sin | Cos | T
Add -> (N + N)
Sub -> (N - N)
Mult -> (N * N)
Pow -> (N ** N)
NConst -> (ConstPos NConst) | Const
Var -> x
Sin -> sin(pi * (N))
Cos -> cos(pi * (N))
Const -> 0 | 1 | 2 | 3 | 4 | 5 | 6 | 7 | 8 | 9
ConstPos -> 1 | 2 | 3 | 4 | 5 | 6 | 7 | 8 | 9
\end{lstlisting}

\paragraph{Finite}

Finite is a simple group where we only create a list of finite numbers. They are created the same way by using previous rules but they are cut at a specific place randomly.

\paragraph{Increasing}
The next three groups are meta-categories, for which no specific grammar is used, but the property is inferred afterwards. If the sequence generated is monotonically increasing, then it is labelled as increasing.

\paragraph{Bounded}

If the sequence is bounded by a number, then it is labelled as bounded. For example $sin(pi*x)$ is bounded.

\paragraph{Unique}

If all numbers in the sequence are distinct, then it is marked as unique. We generate at least first 500 numbers of the sequence to check this property. Of course, this may not be the case for the numbers appearing later in the sequence.

\subsection{Hosting and Maintenance of the Dataset}
The dataset is available through the ETH Research Collection and is directly accessible with DOI 10.3929/ethz-b-000562705 \cite{fact_dataset_through_doi}.
The ETH Research Collection guarantees minimal data retention period of 10 years.
The presence of the dataset will be maintained by the D-ITET DISCO and ISG groups.
Members of DISCO will also be responsible for the maintenance of the dataset in response to queries or any errors found and reported -- please email \href{mailto:fact@ethz.ch}{fact@ethz.ch} to do so.

\section{Baseline Models}
\label{appendix:baselines}
To provide baselines for model performance on the above tasks, we use a total of 24 different models across our bench-marking tasks, namely 4 neural models, 9 classical classifiers, and 11 standard regressors.
The following sections describe the architecture of these models. For consistency, we used the random seed $1234$ wherever it could be specified.

\subsection{Neural Models}
\subsubsection{Dense Neural Network}
The densely connected neural network is composed of three hidden layers with 64, 32 and 16 neurons each, and an input layer with 50 neurons, corresponding to the first 50 integers of the sequence.
Each hidden neuron is activated with the rectified function (ReLU).
The last layer is activated with the sigmoid function in case of classification tasks, and linearly activated in case of regression.
In multi-class-classification, 10 output neurons with sigmoid activation (one for each main class) are used.\\
In terms of hyperparemeter-tuning, we focused on the depth layout and kernel regularisations.
Our grid search contained the following discrete ranges for each parameter: $\text{L1} \in [0.005, 0.01, 0.02] \text{, L2} \in [0.0001, 0.001, 0.01] \text{, depth-layout} \in [(64,32,16,8,4), (64,32,16), (16,16,16)]$ 

\subsubsection{Recurrent Neural Network}
The architecture of our recurrent model consists of three  Long Short-Term Memory \cite{lstm} layers, each with 64, 32 and 16 units respectively.
The first two layers produce the whole sequence, whereas the last layer only yields the last output.
The units of the last layer are fed into a dense-layer with a sigmoidal or linear activation. No dropout was used for RNNs. The hyperparameter-grid-search included: $\text{L1} \in [0, 0.001, 0.01] \text{, L2} \in [0, 0.0001, 0.001] \text{, Dropout} \in [0, 0.1]$

\subsubsection{Convolutional Neural Network}
Here we used a simple stack containing a convolutional layer with kernel size 2, 10 filters and a pooling-layer of size 2, with unit strides.
The network was composed of three of these stacks, terminating in a dense layer with the same activation as above.
The hyperparameter-tuning focused on filter and kernel size: $\text{filters} \in [1,5,10] \text{, kernel size} \in [2,4,6] \text{, stack depth} \in [2,3]$

\subsubsection{Transformer}
We followed the transformer architecture of \cite{transformer} with only the normalisation layers excluded.
We use six transformer blocks for the encoder and three for the decoder decoder.
Each multi-headed attention unit consists of 20 attention heads and the output dimension of the embedding and feed-forward layers is 12.
We focused on the number of attention heads as well as the embedding dimension in our grid search: $\text{heads} \in [5,10,20,40] \text{, embedding dimension} \in [3,6,12,24]$.

\subsection{Standard Classifiers and Regressors}
Table \ref{table:regressor-names} summarizes all standard classifiers and regressors we used.
Each standard model is implemented with the scikit-learn \cite{scikit} library for python. For each model we used its standard parameters. Classification performance is measured in binary-accuracy whereas multiclass performance across all categories is measured with macro-averaged F1-score.

\begin{table}[h!]
\centering
    \begin{tabular}{lr|lr}
    \multicolumn{2}{l}{Classifiers} & \multicolumn{2}{l}{Regressors} \\
    \midrule
    KNNC & k-Nearest Neighbors & KNNR & k-Nearest Neighbors\\
    GNBC & Gaussian naive Bayes & LIR & Linear Regressor\\
    LSVC & Linear Sup. Vector Machine & RIR & Ridge Regressor\\
    DTC & Decision Tree & LAR & Lasso Regressor\\
    RFC & Random Forest & ENR & Elastic Net\\
    GBC & Gradient Boosting & DTR & Decision Tree\\
    ABC & AdaBoost & RFR & Random Forest\\
    XGBC & XG Boost & GBR & Gradient Boosting\\
    DYC & Dummy Classifier & ABR & AdaBoost\\
    & & XGBR & XGBoost\\
    & & DYR & Dummy Regressor\\
    \midrule
    \end{tabular}
\caption{Standard Models}
\label{table:regressor-names}

\end{table}
\begin{table}[h!]
\centering
    \begin{tabular}{lr}
    \midrule
    Max \# of Epochs & 20\\
    Optimizer & Adam\\
    DNN L1 regularizer & 0.001\\
    DNN L2 regularizer & 0.0001\\
    RNN L1 regularizer & 0.001\\
    RNN L2 regularizer & 0.0001\\
    RNN dropout probability & 0 \\
    CNN filter count & 10 \\
    margin distance & 1\\
    \midrule
    \end{tabular}
    \caption{Additional general training hypterparameters}
    \label{table:hyperparameters}
\end{table}

\subsection{Loss and Metrics}
\subsubsection{Binary-Crossentropy}
For our tasks in static mode we used binary-crossentropy as a loss for classification and mean squared logarithmic error for regression.
\subsubsection{Flexible Contrastive Loss}
In dynamic mode  we went for a contrastive loss described in the following formula:
\begin{equation}
    \begin{split}
    d_a & = max(\alpha - d, 0)^2 \\
    d_n & = d^2 \\
    L & = (1 - \lambda) d_a + \lambda d_n
    \end{split}
\end{equation}
where $d$ is the euclidean distance between two sequences in the embedding space and $\alpha$ is the margin-distance which penalises dissimilar pairs only if their distance $d$ is inside its radius.
The goal of this loss is to embed similar sequences near each other in terms of the euclidean distance and different ones further away.
The parameter $\lambda$ functions as a measurement of similarity.
In each dynamic task we define this measurement differently.
In sequence similarity, $\lambda$ is the indicator function between two classes.
In sequence continuation, $\lambda$ is the fraction between the first $n$ similar numbers of two sequences and its total length, whereas in sequence unmasking $\lambda$ is the fraction of masked entries in a sequence paired with its unmasked counterpart.
With this approach we seek to build an embedding space that learns to differentiate different sequences according to our tasks given.

\subsection{Hosting and Maintenance of the Benchmarking Baseline Models}
The bechmarking baseline models are available through the ETH Research Collection and are directly accessible through DOI 10.3929/ethz-b-000565644 \cite{fact_benchmarking_baselines_through_doi}.
The ETH Research Collection guarantees minimal data retention period of 10 years.
The presence of the models will be maintained by the D-ITET DISCO and ISG groups.
Members of DISCO will also be responsible for the maintenance of the models in response to queries or any errors found and reported -- please email \href{mailto:fact@ethz.ch}{fact@ethz.ch} to do so.

\clearpage
\section{Baseline Model Performance}
\label{appendix:baseline-performances}
\begin{table*}[h!]
\scalebox{0.77}{
    \begin{tabular}{l|c|cccccccccc|c}
    
    \toprule
     Model & Dataset & \multicolumn{11}{c}{Scope}\\
    \midrule
    
    \multicolumn{2}{c}{}  & \multicolumn{10}{c}{Within} & Across \\
    \cmidrule(r){3-12}
    \cmidrule(r){13-13}
    
    \multicolumn{2}{c}{} &
    \parbox[t]{2mm}{\multirow{4}{*}{\rotatebox[origin=c]{90}{polynomial}}} &
    \parbox[t]{2mm}{\multirow{4}{*}{\rotatebox[origin=c]{90}{exponential}}} &
    \parbox[t]{2mm}{\multirow{4}{*}{\rotatebox[origin=c]{90}{trigonometric}}} &
    \parbox[t]{2mm}{\multirow{4}{*}{\rotatebox[origin=c]{90}{periodic}}} &
    \parbox[t]{2mm}{\multirow{4}{*}{\rotatebox[origin=c]{90}{finite}}} &
    \parbox[t]{2mm}{\multirow{4}{*}{\rotatebox[origin=c]{90}{modulo}}} &
    \parbox[t]{2mm}{\multirow{4}{*}{\rotatebox[origin=c]{90}{prime}}} &
    \parbox[t]{2mm}{\multirow{4}{*}{\rotatebox[origin=c]{90}{bounded}}} &
    \parbox[t]{2mm}{\multirow{4}{*}{\rotatebox[origin=c]{90}{increasing}}} &
    \parbox[t]{2mm}{\multirow{4}{*}{\rotatebox[origin=c]{90}{unique}}} &
    \parbox[t]{2mm}{\multirow{4}{*}{\rotatebox[origin=c]{90}{all classes}}} \\
    \multicolumn{13}{c}{} \\
    \multicolumn{13}{c}{} \\
    \multicolumn{13}{c}{} \\
    \multicolumn{13}{c}{} \\
    \multicolumn{13}{c}{} \\
    
    \midrule
    \multicolumn{2}{l}{Sequence} & \multicolumn{10}{c}{\multirow{2}{*}{[\textit{binary-accuracy}]}} & \multirow{2}{*}{[\textit{f1-score}]} \\
    \multicolumn{2}{l}{Classification} \\
    \midrule
        \multirow{2}{*}{DNN}         & oeis  & 0.634 &	0.575 &	0.621 &	0.451 &	n.a.  &	0.458 &	0.545 &	0.815 &	0.666 &	0.776 & 0.330 \\
                                     & synth & 0.784 &	0.754 &	0.815 &	0.746 &	0.748 &	0.801 &	0.876 &	0.990 &	0.857 &	0.959 & 0.430 \\
        \cmidrule(r){3-13}
        \multirow{2}{*}{RNN}         & oeis  & 0.588 &	\besto{0.712} &	0.456 &	0.474 &	n.a.  &	0.489 &	0.586 &	\besto{0.860} &	\besto{0.840} &	\besto{0.834} & \besto{0.370} \\
                                     & synth & 0.790 &	0.788 &	0.828 &	 \bests{0.764} &	 \bests{0.755} &	0.825 &	0.907 &	 \bests{0.998} &	 \bests{0.954} &	 \bests{0.976} &  \bests{0.530} \\
        \cmidrule(r){3-13}
        \multirow{2}{*}{CNN}         & oeis  & 0.569 &	0.550 &	0.619 &	0.470 &	n.a.  &	0.483 &	0.514 &	0.800 &	0.575 &	0.694 & 0.220 \\
                                     & synth & 0.769 &	0.704 &	0.792 &	0.727 &	0.736 &	0.769 &	0.843 &	0.976 &	0.752 &	0.793 & 0.390 \\
        \cmidrule(r){3-13}
        \multirow{2}{*}{Transformer} & oeis  & 0.599 &	0.672 &	0.524 &	0.464 &	n.a.  &	0.475 &	0.578 &	0.825 &	0.661 &	0.793 & 0.330 \\
                                     & synth & \bests{0.791} &	0.763 &	0.817 &	0.759 &	0.753 &	0.814 &	0.883 &	0.993 &	0.882 &	0.905 & 0.440 \\
        \cmidrule(r){3-13}
        \multirow{2}{*}{KNNC}        & oeis  & 0.650 &	0.615 &	0.545 &	0.489 &	n.a.  &	0.484 &	0.559 &	0.793 &	0.693 &	0.756 & 0.330 \\
                                     & synth & 0.760 &	0.765 &	0.797 &	0.707 &	0.713 &	0.810 &	0.896 &	0.994 &	0.883 &	0.890 & 0.410 \\
        \cmidrule(r){3-13}
        \multirow{2}{*}{GNBC}        & oeis  & 0.626 &	0.276 &	0.560 &	0.464 &	n.a.  &	0.474 &	0.476 &	0.810 &	0.683 &	0.667 & 0.230 \\
                                     & synth & 0.769 &	0.646 &	0.764 &	0.721 &	0.732 &	0.736 &	0.635 &	0.916 &	0.634 &	0.648 & 0.370 \\
        \cmidrule(r){3-13}
        \multirow{2}{*}{LSVC}        & oeis  & \besto{0.709} &	0.377 &	0.646 &	0.410 &	n.a.  &	0.485 &	0.508 &	0.763 &	0.546 &	0.637 & 0.310 \\
                                     & synth & 0.759 &	0.622 &	0.771 &	0.717 &	0.680 &	0.745 &	0.819 &	0.954 &	0.586 &	0.718 & 0.350 \\
        \cmidrule(r){3-13}
        \multirow{2}{*}{DTC}         & oeis  & 0.618 &	0.607 &	0.497 &	0.480 &	n.a.  &	0.483 &	0.558 &	0.820 &	0.624 &	0.727 & 0.360 \\
                                     & synth & 0.722 &	0.754 &	0.807 &	0.690 &	0.677 &	0.812 &	0.887 &	0.995 &	0.918 &	0.949 & 0.490 \\
        \cmidrule(r){3-13}
        \multirow{2}{*}{RFC}         & oeis  & 0.595 &	0.680 &	0.507 &	0.493 &	n.a.  &	0.468 &	0.563 &	0.843 &	0.579 &	0.787 & 0.340 \\
                                     & synth & 0.789 &	 \bests{0.791} &	 \bests{0.837} &	0.759 &	0.749 &	 \bests{0.830} &	 \bests{0.908} &	 \bests{0.998} &	0.938 &	0.963 & 0.510 \\
        \cmidrule(r){3-13}
        \multirow{2}{*}{GBC}         & oeis  & 0.576 &	0.495 &	0.643 &	0.478 &	n.a.  &	0.470 &	0.548 &	0.821 &	0.623 &	0.788 & 0.270 \\
                                     & synth & 0.785 &	0.746 &	0.804 &	0.758 &	0.751 &	0.803 &	0.872 &	0.990 &	0.835 &	0.860 & 0.400 \\
        \cmidrule(r){3-13}
        \multirow{2}{*}{ABC}         & oeis  & 0.617 &	0.392 &	\besto{0.658} &	0.470 &	n.a.  &	0.457 &	0.497 &	0.782 &	0.511 &	0.691 & 0.310 \\
                                     & synth & 0.773 &	0.676 &	0.777 &	0.741 &	0.737 &	0.761 &	0.829 &	0.958 &	0.669 &	0.766 & 0.380 \\
        \cmidrule(r){3-13}
        \multirow{2}{*}{XGBC}        & oeis  & 0.603 &	0.670 &	0.499 &	0.480 &	n.a.  &	0.477 &	\besto{0.595} &	0.842 &	0.674 &	0.813 & 0.370 \\
                                     & synth & 0.789 &	0.782 &	0.828 &	0.762 &	0.754 &	0.827 &	0.901 &	0.997 &	0.915 &	0.962 & 0.510 \\
        \cmidrule(r){3-13}
        \multirow{2}{*}{DYC}         & oeis  & 0.500 &	0.500 &	0.500 &	\besto{0.500} &	n.a.  &	\besto{0.500} &	0.500 &	0.500 &	0.500 &	0.500 & 0.500 \\
                                     & synth & 0.500 &	0.500 &	0.500 &	0.500 &	0.500 &	0.500 &	0.500 &	0.500 &	0.500 &	0.500 & 0.500 \\
    \midrule
    \multicolumn{2}{l}{Next Sequence-Part} & \multicolumn{11}{c}{\multirow{2}{*}{[\textit{binary-accuracy}]}} \\
    \multicolumn{2}{l}{Prediction} \\
    \midrule
        \multirow{2}{*}{DNN}         & oeis  & 0.658 & 0.664 &	0.719 &	0.778 &	n.a.	& 0.760 & 0.744 & 0.749 & 0.753 & 0.726 & 0.733 \\
                                     & synth & 0.941 & 0.927 &	0.940 &	0.943 &	0.939	& 0.918 & 0.914 & 0.943 & 0.924 & 0.936 & 0.943 \\
        \cmidrule(r){3-13}
        \multirow{2}{*}{RNN}         & oeis  & \besto{0.869} & \besto{0.890} &	\besto{0.827} &	\besto{0.849} &	n.a.	& \besto{0.855} & \besto{0.860} & \besto{0.833} & \besto{0.876} & \besto{0.889} & \besto{0.869} \\
                                     & synth &  \bests{0.988} &  \bests{0.973} &	 \bests{0.979} &	 \bests{0.987} &	 \bests{0.988}	&  \bests{0.972} &  \bests{0.955} &  \bests{0.978} &  \bests{0.982} &  \bests{0.976} &  \bests{0.984} \\
        \cmidrule(r){3-13}
        \multirow{2}{*}{CNN}         & oeis  & 0.526 & 0.520 &	0.540 &	0.566 &	n.a.	& 0.548 & 0.535 & 0.566 & 0.547 & 0.539 & 0.551 \\
                                     & synth & 0.893 & 0.912 &	0.922 &	0.903 &	0.895	& 0.898 & 0.890 & 0.915 & 0.885 & 0.901 & 0.900 \\
        \cmidrule(r){3-13}
        \multirow{2}{*}{Transformer} & oeis  & 0.690 & 0.666 &	0.707 &	0.792 &	n.a.	& 0.759 & 0.747 & 0.752 & 0.744 & 0.737 & 0.736 \\
                                     & synth & 0.947 & 0.930 &	0.945 &	0.947 &	0.949	& 0.927 & 0.919 & 0.946 & 0.926 & 0.943 & 0.938 \\
    \bottomrule 
    
    \end{tabular}
}
\caption{
    The results for the classification and next sequence-part predicition, evaluated both within categories and across the whole dataset.
    MLP, RNN, and CNN stand for multi-layer perceptron, recurrent neural network, and convolutional neural network.
    \besto{Emphasis} and \bests{emphasis} mark the best performing models for the OEIS and synthetic data, respectively. For both F1 score and binary accuracy, higher is better.
}
\label{tab:results-classification-nspp}
\end{table*}

\begin{table*}[h!]
    \centering
\scalebox{0.77}{
    \begin{tabular}{l|c|c|cccccccccc}
    
    \toprule
     Model & Dataset & Metric & \multicolumn{10}{c}{Scope}\\
    \midrule
    \multicolumn{3}{c}{}  & \multicolumn{10}{c}{Within} \\
    \cmidrule(r){4-13}
    \multicolumn{3}{c}{} &
    \parbox[t]{2mm}{\multirow{4}{*}{\rotatebox[origin=c]{90}{polynomial}}} &
    \parbox[t]{2mm}{\multirow{4}{*}{\rotatebox[origin=c]{90}{exponential}}} &
    \parbox[t]{2mm}{\multirow{4}{*}{\rotatebox[origin=c]{90}{trigonometric}}} &
    \parbox[t]{2mm}{\multirow{4}{*}{\rotatebox[origin=c]{90}{periodic}}} &
    \parbox[t]{2mm}{\multirow{4}{*}{\rotatebox[origin=c]{90}{finite}}} &
    \parbox[t]{2mm}{\multirow{4}{*}{\rotatebox[origin=c]{90}{modulo}}} &
    \parbox[t]{2mm}{\multirow{4}{*}{\rotatebox[origin=c]{90}{prime}}} &
    \parbox[t]{2mm}{\multirow{4}{*}{\rotatebox[origin=c]{90}{bounded}}} &
    \parbox[t]{2mm}{\multirow{4}{*}{\rotatebox[origin=c]{90}{increasing}}} &
    \parbox[t]{2mm}{\multirow{4}{*}{\rotatebox[origin=c]{90}{unique}}} \\
    \multicolumn{13}{c}{} \\
    \multicolumn{13}{c}{} \\
    \multicolumn{13}{c}{} \\
    \multicolumn{13}{c}{} \\
    \multicolumn{13}{c}{} \\
    
    \midrule
    \multicolumn{3}{l}{Sequence} & \multicolumn{10}{c}{\multirow{2}{*}{[\textit{binary-accuracy}]}} \\
    \multicolumn{3}{l}{Similarity} \\
    \midrule
        \multirow{6}{*}{DNN}         & \multirow{3}{*}{oeis}  & Top-1 & \besto{0.20} & \besto{0.00} & 0.08 & 0.16 & n.a. & 0.09 & 0.12 & 0.07 & \besto{0.09} & 0.08 \\
                                     &                        & Top-3 & 0.23 & 0.00 & 0.19 & 0.16 & n.a. & 0.23 & 0.21 & 0.29 & \besto{0.32} & 0.26 \\
                                     &                        & Top-5 & 0.44 & 0.00 & 0.27 & 0.33 & n.a. & 0.30 & 0.34 & 0.43 & 0.45 & 0.48 \\
        \cmidrule(r){4-13}
                                     & \multirow{3}{*}{synth} & Top-1 & 0.12 & 0.13 & 0.19 & 0.10 & 0.09 & 0.11 & 0.32 & 0.10 & 0.07 & 0.09 \\
                                     &                        & Top-3 & 0.27 & 0.34 & 0.30 & 0.25 & 0.31 & 0.23 & 0.54 & 0.38 & 0.27 & 0.33 \\
                                     &                        & Top-5 & 0.43 & 0.46 & 0.44 & 0.44 & 0.40 &  \bests{0.57} & 0.69 & 0.57 &  \bests{0.44} & 0.48 \\
        \cmidrule(r){4-13}
        \multirow{6}{*}{RNN}         & \multirow{3}{*}{oeis}  & Top-1 & 0.07 & \besto{0.00} & 0.04 & \besto{0.20} & n.a. & \besto{0.13} & 0.14 & 0.06 & 0.06 & 0.11 \\
                                     &                        & Top-3 & 0.25 & 0.00 & 0.27 & \besto{0.50} & n.a. & 0.35 & 0.16 & 0.34 & 0.31 & 0.25 \\
                                     &                        & Top-5 & \besto{0.53} & 0.00 & 0.40 & \besto{0.50} & n.a. & 0.41 & 0.34 & 0.45 & 0.44 & 0.42 \\
        \cmidrule(r){4-13}
                                     & \multirow{3}{*}{synth} & Top-1 & 0.17 & 0.11 & 0.20 & 0.08 & 0.10 &  \bests{0.13} & 0.35 & 0.12 & 0.07 & 0.08 \\
                                     &                        & Top-3 & 0.32 & 0.29 & 0.39 & 0.32 & 0.32 & 0.25 & 0.56 & 0.40 & 0.29 & 0.32 \\
                                     &                        & Top-5 & 0.41 & 0.38 & 0.47 & 0.42 & 0.47 & 0.49 &  \bests{0.70} & 0.56 & 0.41 & 0.51 \\
        \cmidrule(r){4-13}
        \multirow{6}{*}{CNN}         & \multirow{3}{*}{oeis}  & Top-1 & 0.15 & \besto{0.00} & 0.06 & 0.00 & n.a. & 0.06 & 0.16 & \besto{0.14} & 0.07 & \besto{0.15} \\
                                     &                        & Top-3 & \besto{0.34} & \besto{0.12} & 0.31 & 0.00 & n.a. & \besto{0.38} & \besto{0.32} & \besto{0.43} & 0.26 & 0.36 \\
                                     &                        & Top-5 & 0.34 & \besto{0.75} & 0.41 & 0.40 & n.a. & 0.47 & \besto{0.46} & \besto{0.59} & 0.39 & 0.50 \\
        \cmidrule(r){4-13}
                                     & \multirow{3}{*}{synth} & Top-1 & 0.09 &  \bests{0.14} & 0.16 & 0.07 & 0.13 & 0.11 & 0.32 &  \bests{0.17} &  \bests{0.08} &  \bests{0.10} \\
                                     &                        & Top-3 & 0.37 &  \bests{0.39} & 0.37 &  \bests{0.42} & 0.29 &  \bests{0.32} &  \bests{0.61} & 0.49 & 0.30 &  \bests{0.39} \\
                                     &                        & Top-5 & 0.50 & 0.41 & 0.60 & 0.48 & 0.47 & 0.51 & 0.66 & 0.64 &  \bests{0.44} & 0.60 \\
        \cmidrule(r){4-13}
        \multirow{6}{*}{Transformer} & \multirow{3}{*}{oeis}  & Top-1 & 0.03 & \besto{0.00} & \besto{0.13} & 0.00 & n.a. & 0.05 & \besto{0.18} & 0.11 & \besto{0.09} & 0.09 \\
                                     &                        & Top-3 & 0.31 & 0.00 & \besto{0.39} & 0.10 & n.a. & 0.33 & 0.28 & 0.42 & 0.29 & \besto{0.38} \\
                                     &                        & Top-5 & 0.48 & 0.66 & \besto{0.43} & 0.20 & n.a. & \besto{0.49} & 0.40 & 0.50 & 0.43 & \besto{0.55} \\
        \cmidrule(r){4-13}
                                     & \multirow{3}{*}{synth} & Top-1 &  \bests{0.21} & 0.13 &  \bests{0.31} &  \bests{0.12} &  \bests{0.17} & 0.10 &  \bests{0.36} & 0.13 & 0.07 &  \bests{0.10} \\
                                     &                        & Top-3 &  \bests{0.44} & 0.32 &  \bests{0.41} & 0.41 &  \bests{0.37} &  \bests{0.32} & 0.56 &  \bests{0.50} &  \bests{0.32} & 0.35 \\
                                     &                        & Top-5 &  \bests{0.67} &  \bests{0.52} &  \bests{0.63} &  \bests{0.51} &  \bests{0.52} & 0.55 & 0.65 &  \bests{0.61} &  \bests{0.44} &  \bests{0.58} \\
    \bottomrule
    
    \end{tabular}
}
\caption{
    The accuracy results for the sequence similarity task, evaluated both within categories and across the whole dataset.
    \besto{Emphasis} and \bests{emphasis} mark the best performing models for the OEIS and synthetic data, respectively.
    For binary accuracy, higher is better.
}
\label{tab:results-similarity-accuracy}
\end{table*}

\begin{table*}[h!]
\centering
\scalebox{0.77}{
    \begin{tabular}{l|c|cccccccccc|c}
    
    \toprule
     Model & Dataset & \multicolumn{11}{c}{Scope}\\
    \midrule
    
    \multicolumn{2}{c}{}  & \multicolumn{10}{c}{Within} & Across \\
    \cmidrule(r){3-12}
    \cmidrule(r){13-13}
    
    \multicolumn{2}{c}{} &
    \parbox[t]{2mm}{\multirow{4}{*}{\rotatebox[origin=c]{90}{polynomial}}} &
    \parbox[t]{2mm}{\multirow{4}{*}{\rotatebox[origin=c]{90}{exponential}}} &
    \parbox[t]{2mm}{\multirow{4}{*}{\rotatebox[origin=c]{90}{trigonometric}}} &
    \parbox[t]{2mm}{\multirow{4}{*}{\rotatebox[origin=c]{90}{periodic}}} &
    \parbox[t]{2mm}{\multirow{4}{*}{\rotatebox[origin=c]{90}{finite}}} &
    \parbox[t]{2mm}{\multirow{4}{*}{\rotatebox[origin=c]{90}{modulo}}} &
    \parbox[t]{2mm}{\multirow{4}{*}{\rotatebox[origin=c]{90}{prime}}} &
    \parbox[t]{2mm}{\multirow{4}{*}{\rotatebox[origin=c]{90}{bounded}}} &
    \parbox[t]{2mm}{\multirow{4}{*}{\rotatebox[origin=c]{90}{increasing}}} &
    \parbox[t]{2mm}{\multirow{4}{*}{\rotatebox[origin=c]{90}{unique}}} &
    \parbox[t]{2mm}{\multirow{4}{*}{\rotatebox[origin=c]{90}{all classes}}} \\
    \multicolumn{13}{c}{} \\
    \multicolumn{13}{c}{} \\
    \multicolumn{13}{c}{} \\
    \multicolumn{13}{c}{} \\
    \multicolumn{13}{c}{} \\
    
    \midrule
    \multicolumn{2}{l}{Sequence} & \multicolumn{11}{c}{\multirow{2}{*}{[\textit{root-mean-squared-log-error}]}} \\
    \multicolumn{2}{l}{Continuation} \\
    \midrule
        \multirow{2}{*}{DNN}         & oeis  &         0.750 &         0.700 &         0.588 &         0.576 & n.a.	         &         0.567 &         0.617 &         0.519 &         0.600 &         0.614 &         0.597 \\
                                     & synth &         0.496 &         0.408 &         0.345 &         0.485 & 0.489         &         0.398 &         0.379 &         0.372 &         0.477 &         0.452 &         0.430 \\
        \cmidrule(r){3-13}
        \multirow{2}{*}{RNN}         & oeis  & \besto{0.738} & \besto{0.692} & \besto{0.561} & \besto{0.561} & n.a.	         &         0.554 &         0.602 &         0.506 &         0.577 &         0.614 &         0.603 \\
                                     & synth & \bests{0.470} & \bests{0.375} &         0.317 & \bests{0.466} & 0.461         &         0.381 &         0.345 &         0.351 &         0.457 &         0.424 &         0.406 \\
        \cmidrule(r){3-13}
        \multirow{2}{*}{CNN}         & oeis  &         0.776 &         0.768 &         0.765 &         0.679 & n.a.	         &         0.727 &         0.758 &         0.686 &         0.730 &         0.737 &         0.733 \\
                                     & synth &         0.586 &         0.550 &         0.498 &         0.585 & 0.581         &         0.557 &         0.623 &         0.536 &         0.599 &         0.612 &         0.579 \\
        \cmidrule(r){3-13}
        \multirow{2}{*}{Transformer} & oeis  &         1.632 &         1.596 &         0.573 &         1.113 & n.a.	         & \besto{0.545} & \besto{0.573} & \besto{0.503} & \besto{0.575} & \besto{0.593} & \besto{0.578} \\
                                     & synth &         2.051 &         1.420 & \bests{0.308} &         1.978 & \bests{0.452} & \bests{0.365} & \bests{0.335} & \bests{0.341} & \bests{0.449} & \bests{0.415} & \bests{0.395} \\
        \cmidrule(r){3-13}
        \multirow{2}{*}{KNNR}        & oeis  &         0.955 &         0.874 &         0.761 &         0.807 & n.a.	         &         0.730 &         0.796 &         0.669 &         0.783 &         0.832 &         0.808 \\
                                     & synth &         0.575 &         0.451 &         0.373 &         0.560 & 0.551         &         0.459 &         0.401 &         0.433 &         0.564 &         0.513 &         0.486 \\
        \cmidrule(r){3-13}
        \multirow{2}{*}{LIR}         & oeis  &         0.872 &         0.784 &         0.723 &         0.880 & n.a.	         &         0.710 &         0.846 &         0.704 &         0.786 &         0.821 &         0.797 \\
                                     & synth &         0.694 &         0.633 &         0.545 &         0.696 & 0.692         &         0.611 &         0.770 &         0.613 &         0.701 &         0.724 &         0.682 \\
        \cmidrule(r){3-13}
        \multirow{2}{*}{RIR}         & oeis  &         0.873 &         0.784 &         0.721 &         0.875 & n.a.	         &         0.713 &         0.846 &         0.703 &         0.786 &         0.822 &         0.797 \\
                                     & synth &         0.692 &         0.632 &         0.546 &         0.695 & 0.692         &         0.613 &         0.769 &         0.613 &         0.701 &         0.725 &         0.682 \\
        \cmidrule(r){3-13}
        \multirow{2}{*}{LAR}         & oeis  &         0.910 &         0.750 &         0.727 &         1.012 & n.a.	         &         0.734 &         0.882 &         0.743 &         0.798 &         0.856 &         0.827 \\
                                     & synth &         0.750 &         0.703 &         0.615 &         0.754 & 0.748         &         0.688 &         0.812 &         0.683 &         0.765 &         0.783 &         0.747 \\
        \cmidrule(r){3-13}
        \multirow{2}{*}{ENR}         & oeis  &         0.886 &         0.756 &         0.722 &         0.951 & n.a.	         &         0.723 &         0.862 &         0.724 &         0.793 &         0.840 &         0.814 \\
                                     & synth &         0.722 &         0.672 &         0.583 &         0.727 & 0.723         &         0.656 &         0.794 &         0.651 &         0.734 &         0.754 &         0.716 \\
        \cmidrule(r){3-13}
        \multirow{2}{*}{DTR}         & oeis  &         0.868 &         0.801 &         0.693 &         0.741 & n.a.	         &         0.663 &         0.731 &         0.618 &         0.695 &         0.749 &         0.730 \\
                                     & synth &         0.496 &         0.392 &         0.328 &         0.492 & 0.487         &         0.391 &         0.351 &         0.368 &         0.490 &         0.445 &         0.427 \\
        \cmidrule(r){3-13}
        \multirow{2}{*}{RFR}         & oeis  &         0.871 &         0.797 &         0.696 &         0.740 & n.a.	         &         0.666 &         0.730 &         0.619 &         0.696 &         0.748 &         0.730 \\
                                     & synth &         0.496 &         0.393 &         0.325 &         0.492 & 0.488         &         0.396 &         0.348 &         0.368 &         0.491 &         0.446 &         0.427 \\
        \cmidrule(r){3-13}
        \multirow{2}{*}{GBR}         & oeis  &         0.857 &         0.789 &         0.622 &         0.694 & n.a.	         &         0.650 &         0.706 &         0.578 &         0.694 &         0.726 &         0.702 \\
                                     & synth &         0.544 &         0.459 &         0.377 &         0.540 & 0.535         &         0.446 &         0.420 &         0.420 &         0.545 &         0.510 &         0.484 \\
        \cmidrule(r){3-13}
        \multirow{2}{*}{ABR}         & oeis  &         0.907 &         0.868 &         0.782 &         0.894 & n.a.	         &         0.754 &         0.837 &         0.776 &         0.796 &         0.878 &         0.842 \\
                                     & synth &         0.635 &         0.621 &         0.542 &         0.667 & 0.659         &         0.594 &         0.652 &         0.635 &         0.658 &         0.672 &         0.662 \\
        \cmidrule(r){3-13}
        \multirow{2}{*}{XGBR}        & oeis  &         0.869 &         0.796 &         0.679 &         0.726 & n.a.	         &         0.651 &         0.707 &         0.607 &         0.688 &         0.735 &         0.719 \\
                                     & synth &         0.499 &         0.400 &         0.330 &         0.497 & 0.490         &         0.400 &         0.360 &         0.372 &         0.496 &         0.452 &         0.433 \\
        \cmidrule(r){3-13}
        \multirow{2}{*}{DYR}         & oeis  &         0.972 &         0.770 &         0.821 &         1.202 & n.a.	         &         0.788 &         0.912 &         0.883 &         0.860 &         0.930 &         0.923 \\
                                     & synth &         0.832 &         0.868 &         0.807 &         0.847 & 0.844         &         0.858 &         0.877 &         0.866 &         0.876 &         0.871 &         0.877 \\

    \bottomrule 
    
    \end{tabular}
}
\caption{
    The results for the sequence continuation task, evaluated both within categories and across the whole dataset.
    \besto{Emphasis} and \bests{emphasis} mark the best performing models for the OEIS and synthetic data, respectively.
    For RMSLE, lower is better.
}
\label{tab:results-continuation}
\end{table*}

\begin{table*}[h!]
\centering
\scalebox{0.77}{
    \begin{tabular}{l|c|c|cccccccccc|c}
    \toprule
    
     Model & Dataset & Metric & \multicolumn{11}{c}{Scope}\\
    \midrule
    
    \multicolumn{3}{c}{}  & \multicolumn{10}{c}{Within} & Across \\
    \cmidrule(r){4-13}
    \cmidrule(r){14-14}
    
    \multicolumn{3}{c}{} &
    \parbox[t]{2mm}{\multirow{4}{*}{\rotatebox[origin=c]{90}{polynomial}}} &
    \parbox[t]{2mm}{\multirow{4}{*}{\rotatebox[origin=c]{90}{exponential}}} &
    \parbox[t]{2mm}{\multirow{4}{*}{\rotatebox[origin=c]{90}{trigonometric}}} &
    \parbox[t]{2mm}{\multirow{4}{*}{\rotatebox[origin=c]{90}{periodic}}} &
    \parbox[t]{2mm}{\multirow{4}{*}{\rotatebox[origin=c]{90}{finite}}} &
    \parbox[t]{2mm}{\multirow{4}{*}{\rotatebox[origin=c]{90}{modulo}}} &
    \parbox[t]{2mm}{\multirow{4}{*}{\rotatebox[origin=c]{90}{prime}}} &
    \parbox[t]{2mm}{\multirow{4}{*}{\rotatebox[origin=c]{90}{bounded}}} &
    \parbox[t]{2mm}{\multirow{4}{*}{\rotatebox[origin=c]{90}{increasing}}} &
    \parbox[t]{2mm}{\multirow{4}{*}{\rotatebox[origin=c]{90}{unique}}} &
    \parbox[t]{2mm}{\multirow{4}{*}{\rotatebox[origin=c]{90}{all classes}}} \\
    \multicolumn{14}{c}{} \\
    \multicolumn{14}{c}{} \\
    \multicolumn{14}{c}{} \\
    \multicolumn{14}{c}{} \\
    \multicolumn{14}{c}{} \\
    
    \midrule
    \multicolumn{3}{l}{Sequence} & \multicolumn{11}{c}{\multirow{2}{*}{[\textit{top-k-root-mean-squared-error}]}} \\
    \multicolumn{3}{l}{Similarity} \\
    \midrule
        \multirow{6}{*}{DNN}         & \multirow{3}{*}{oeis}  & Top-1 & 2.037 & 1.662 & 3.542 & 1.392 & n.a.    & 1.867 & 1.432 & 0.850 & 2.488 & 1.300 & 2.668 \\
                                     &                        & Top-3 & 0.931 & 1.662 & 0.442 & 0.390 & n.a.    & 0.785 & 0.587 & 0.367 & 0.900 & 0.912 & 0.599 \\
                                     &                        & Top-5 & 0.855 & 1.662 & 0.361 & \besto{0.152} & n.a.    & 0.696 & 0.433 & \besto{0.250} & 0.561 & 0.912 & 0.301 \\
        \cmidrule(r){4-14}
                                     & \multirow{3}{*}{synth} & Top-1 & 5.847 & 3.261 & 5.616 & 2.790 & 2.695   & 1.597 & 1.896 & 3.951 & 2.153 & 2.302 & 2.144 \\
                                     &                        & Top-3 & 0.870 & 0.927 & 0.365 & 1.192 & 1.483   & 0.475 & 1.246 & 0.534 & 1.130 & 1.900 & 1.690 \\
                                     &                        & Top-5 & 0.752 & 0.549 & 0.365 & 0.766 & 1.483   & 0.448 & 0.944 & 0.534 & 0.825 & 0.520 & 1.690 \\
        \cmidrule(r){4-14}
        \multirow{6}{*}{RNN}         & \multirow{3}{*}{oeis}  & Top-1 & 2.089 & 2.278 & 1.624 & 1.735 & n.a.    & 1.326 & 1.831 & 1.264 & 1.473 & 1.450 & 1.367 \\
                                     &                        & Top-3 & 1.038 & 1.025 & 0.677 & 0.661 & n.a.    & 0.645 & 0.823 & 0.597 & 0.617 & 0.658 & 0.596 \\
                                     &                        & Top-5 & 0.706 & 0.665 & 0.429 & 0.439 & n.a.    & 0.435 & 0.585 & 0.406 & 0.427 & 0.457 & 0.383 \\
        \cmidrule(r){4-14}
                                     & \multirow{3}{*}{synth} & Top-1 & 1.889 & 1.488 & 0.974 & 2.193 & 2.331   & 1.034 & 1.210 & 1.290 & 1.453 & 1.604 & 1.509 \\
                                     &                        & Top-3 & 0.784 & 0.566 & 0.405 & 0.954 & 1.033   & 0.448 & 0.508 & 0.589 & 0.654 & 0.713 & 0.667 \\
                                     &                        & Top-5 & 0.506 & 0.364 & 0.244 & 0.595 & 0.607   & 0.290 & 0.320 & 0.355 & 0.401 & 0.476 & 0.438 \\
        \cmidrule(r){4-14}
        \multirow{6}{*}{CNN}         & \multirow{3}{*}{oeis}  & Top-1 & 3.117 & 2.943 & 1.957 & 1.715 & n.a.    & 2.426 & 1.570 & 1.430 & 1.950 & 1.490 & 1.825 \\
                                     &                        & Top-3 & 1.309 & 1.302 & 0.679 & 0.813 & n.a.    & 1.046 & 0.702 & 0.501 & 0.579 & 1.003 & 0.807 \\
                                     &                        & Top-5 & 0.631 & 0.646 & 0.585 & 0.521 & n.a.    & 0.736 & \besto{0.361} & 0.296 & 0.406 & 0.711 & 0.428 \\
        \cmidrule(r){4-14}
                                     & \multirow{3}{*}{synth} & Top-1 & 2.475 & 1.662 & 1.399 & 3.533 & 2.338   & 1.848 & 7.013 & 2.288 & 2.333 & 2.033 & 2.348 \\
                                     &                        & Top-3 & 1.060 & 0.695 & 1.115 & 1.031 & 1.151   & 0.880 & 1.007 & 1.152 & 0.826 & 0.871 & 1.105 \\
                                     &                        & Top-5 & 0.664 & 0.440 & 0.836 & 0.777 & 0.697   & 0.647 & 0.648 & 0.682 & 0.463 & 0.613 & 0.643 \\
        \cmidrule(r){4-14}
        \multirow{6}{*}{Transformer} & \multirow{3}{*}{oeis}  & Top-1 & \besto{1.503} & \besto{1.408} & \besto{0.984} & \besto{0.786} & n.a.    & \besto{0.977} & \besto{1.161} & \besto{0.663} & \besto{1.022} & \besto{0.935} & \besto{0.847} \\
                                     &                        & Top-3 & \besto{0.746} & \besto{0.748} & \besto{0.398} & \besto{0.382} & n.a.    & \besto{0.454} & \besto{0.584} & \besto{0.348} & \besto{0.499} & \besto{0.438} & \besto{0.383} \\
                                     &                        & Top-5 & \besto{0.529} & \besto{0.524} & \besto{0.253} & 0.287 & n.a.    & \besto{0.285} & 0.414 & 0.284 & \besto{0.376} & \besto{0.288} & \besto{0.267} \\
        \cmidrule(r){4-14}
                                     & \multirow{3}{*}{synth} & Top-1 &  \bests{1.816} &  \bests{0.962} &  \bests{0.788} &  \bests{1.730} &  \bests{1.651}   &  \bests{0.811} &  \bests{0.801} &  \bests{0.873} &  \bests{1.084} &  \bests{1.198} &  \bests{1.258} \\
                                     &                        & Top-3 &  \bests{0.727} &  \bests{0.410} &  \bests{0.312} &  \bests{0.787} &  \bests{0.706}   &  \bests{0.316} &  \bests{0.331} &  \bests{0.389} &  \bests{0.447} &  \bests{0.507} &  \bests{0.490} \\
                                     &                        & Top-5 &  \bests{0.448} &  \bests{0.248} &  \bests{0.202} &  \bests{0.484} &  \bests{0.418}   &  \bests{0.225} &  \bests{0.205} &  \bests{0.233} &  \bests{0.270} &  \bests{0.331} &  \bests{0.284} \\
    \midrule
    \multicolumn{3}{l}{Sequence} & \multicolumn{11}{c}{\multirow{2}{*}{[\textit{top-k-root-mean-squared-error}]}} \\
    \multicolumn{3}{l}{Unmasking} \\
    \midrule
        \multirow{6}{*}{DNN}         & \multirow{3}{*}{oeis}  & Top-1 & 3.702 & \besto{3.529} & \besto{3.460} & 3.248 & n.a.    & 3.451 & \besto{3.451} & 3.274 & \besto{3.307} & 3.315 & 3.384 \\
                                     &                        & Top-3 & 3.305 & 3.163 & 2.936 & 2.917 & n.a.    & 3.114 & 3.059 & 2.976 & 3.061 & 3.027 & 3.061 \\
                                     &                        & Top-5 & 3.125 & 3.000 & 2.748 & 2.779 & n.a.    & 2.972 & 2.880 & 2.878 & 2.925 & 2.903 & 2.918 \\
        \cmidrule(r){4-14}
                                     & \multirow{3}{*}{synth} & Top-1 & 4.240 & 3.711 & 3.374 & 4.062 & 4.092   & 3.633 & 3.786 & 3.619 & 3.839 & 3.766 & 3.855 \\
                                     &                        & Top-3 & 3.898 & 3.448 & 2.958 & 3.710 & 3.744   & 3.353 & 3.321 & 3.356 & 3.547 & 3.470 & 3.524 \\
                                     &                        & Top-5 & 3.776 & 3.335 & 2.825 & 3.582 & 3.593   & 3.239 & 3.161 & 3.236 & 3.441 & 3.361 & 3.408 \\
        \cmidrule(r){4-14}
        \multirow{6}{*}{RNN}         & \multirow{3}{*}{oeis}  & Top-1 & 3.789 & 3.720 & 3.548 & 3.314 & n.a.    & 3.484 & 3.499 & 3.320 & 3.490 & 3.485 & 3.455 \\
                                     &                        & Top-3 & 3.206 & 3.175 & 3.052 & 2.876 & n.a.    & 3.060 & 2.998 & 2.908 & 3.109 & 3.083 & 3.091 \\
                                     &                        & Top-5 & 3.010 & 2.990 & 2.847 & 2.688 & n.a.    & 2.898 & 2.830 & 2.757 & 2.925 & 2.909 & 2.944 \\
        \cmidrule(r){4-14}
                                     & \multirow{3}{*}{synth} & Top-1 & 4.141 & 3.765 & 3.368 & 4.094 & 4.115   & 3.535 & 3.775 & 3.696 & 3.674 & 3.913 & 3.855 \\
                                     &                        & Top-3 & 3.663 & 3.463 & 3.082 & 3.639 & 3.663   & 3.254 & 3.314 & 3.346 & 3.328 & 3.515 & 3.511 \\
                                     &                        & Top-5 & 3.473 & 3.291 & 2.961 & 3.472 & 3.507   & 3.140 & 3.162 & 3.197 & 3.208 & 3.370 & 3.379 \\
        \cmidrule(r){4-14}
        \multirow{6}{*}{CNN}         & \multirow{3}{*}{oeis}  & Top-1 & 3.738 & 3.699 & 3.615 & 3.268 & n.a.    & \besto{3.383} & 3.521 & \besto{3.165} & 3.453 & \besto{3.257} & \besto{3.355} \\
                                     &                        & Top-3 & \besto{2.943} & \besto{2.922} & \besto{2.873} & \besto{2.594} & n.a.    & \besto{2.683} & \besto{2.785} & \besto{2.510} & \besto{2.727} & \besto{2.643} & \besto{2.690} \\
                                     &                        & Top-5 & \besto{2.689} & \besto{2.631} & \besto{2.577} & \besto{2.370} & n.a.    & \besto{2.437} & \besto{2.539} & \besto{2.260} & \besto{2.490} & \besto{2.423} & \besto{2.440} \\
        \cmidrule(r){4-14}
                                     & \multirow{3}{*}{synth} & Top-1 &  \bests{3.906} &  \bests{3.584} &  \bests{3.223} &  \bests{3.791} &  \bests{3.886}   & 3.500 & 3.646 &  \bests{3.531} &  \bests{3.627} &  \bests{3.702} &  \bests{3.611} \\
                                     &                        & Top-3 &  \bests{3.179} &  \bests{3.100} &  \bests{2.811} &  \bests{3.128} &  \bests{3.168}   &  \bests{2.968} &  \bests{3.112} &  \bests{3.008} &  \bests{2.988} &  \bests{3.122} &  \bests{3.033} \\
                                     &                        & Top-5 &  \bests{2.931} &  \bests{2.799} &  \bests{2.642} &  \bests{2.891} &  \bests{2.898}   &  \bests{2.784} &  \bests{2.890} &  \bests{2.834} &  \bests{2.747} &  \bests{2.899} &  \bests{2.812} \\
        \cmidrule(r){4-14}
        \multirow{6}{*}{Transformer} & \multirow{3}{*}{oeis}  & Top-1 & \besto{3.635} & 3.674 & 3.586 & \besto{3.245} & n.a.    & 3.465 & 3.556 & 3.378 & 3.456 & 3.606 & 3.524 \\
                                     &                        & Top-3 & 3.042 & 3.045 & 2.951 & 2.798 & n.a.    & 2.991 & 2.974 & 2.903 & 2.947 & 3.038 & 3.017 \\
                                     &                        & Top-5 & 2.820 & 2.781 & 2.658 & 2.606 & n.a.    & 2.808 & 2.717 & 2.717 & 2.756 & 2.816 & 2.811 \\
        \cmidrule(r){4-14}
                                     & \multirow{3}{*}{synth} & Top-1 & 3.953 & 3.719 & 3.403 & 3.968 & 4.079   &  \bests{3.478} &  \bests{3.638} & 3.605 & 3.717 & 3.780 & 3.757 \\
                                     &                        & Top-3 & 3.374 & 3.259 & 3.019 & 3.395 & 3.475   & 3.190 & 3.115 & 3.226 & 3.293 & 3.359 & 3.291 \\
                                     &                        & Top-5 & 3.171 & 3.067 & 2.872 & 3.161 & 3.259   & 3.041 & 2.941 & 3.045 & 3.121 & 3.173 & 3.091 \\
    \bottomrule 
    
    \end{tabular}
}
\caption{
    The top-$k$ RMSE results for the sequence similarity and unmasking tasks, evaluated both within categories and across the whole dataset.
    \besto{Emphasis} and \bests{emphasis} mark the best performing models for the OEIS and synthetic data, respectively.
    The masking probability is $0.25$.
    For top-$k$-RMSE, lower is better.
}
\label{tab:results-similarity-unmasking-rmse}
\end{table*}

\clearpage

\subsection{Compute Time Breakdown}
\label{appendix:compute-time}

\renewcommand{\arraystretch}{1.25}

\begin{table*}[!ht]
    \centering
    \begin{tabular}{l|ccccc}
    
    \toprule
    Model & \multicolumn{5}{c}{Task}\\
    \midrule
    
    
    \multicolumn{1}{c}{} &
    \parbox[t]{12mm}{\multirow{4}{*}{\centering\rotatebox[origin=c]{60}{classification}}} &
    \parbox[t]{12mm}{\multirow{4}{*}{\centering\rotatebox[origin=c]{60}{next part pred.}}} &
    \parbox[t]{12mm}{\multirow{4}{*}{\centering\rotatebox[origin=c]{60}{continuation}}} &
    \parbox[t]{12mm}{\multirow{4}{*}{\centering\rotatebox[origin=c]{60}{similarity}}} &
    \parbox[t]{12mm}{\multirow{4}{*}{\centering\rotatebox[origin=c]{60}{unmasking}}} \\
    \multicolumn{6}{c}{} \\
    \multicolumn{6}{c}{} \\
    \multicolumn{6}{c}{} \\
    \multicolumn{6}{c}{} \\

    \midrule
    \multicolumn{1}{c}{} & \multicolumn{5}{c}{\textit{mean task training and evaluation time in minutes}} \\
    \midrule
        
        DNN & 23 & 214 & 98 & 32 & 168 \\
        RNN & 44 & 428 & 234 & 57 & 345 \\
        CNN & 35 & 388 & 198 & 48 & 246 \\
        Transformer & 48 & 465 & 256 & 45 & 326 \\
        KNNC & 67 & - & - & - & - \\
        GNBC & 7 & - & - & - & - \\
        DTC & 14 & - & - & - & - \\
        XGBC & 17 & - & - & - & - \\
        LSVC & 26 & - & - & - & - \\
        RFC & 14 & - & - & - & - \\
        GBC & 9 & - & - & - & - \\
        ABC & 23 & - & - & - & - \\
        XGBC & 11 & - & - & - & - \\
        KNNR & - & - & - & 72 & - \\
        RIR & - & - & - & 42 & - \\
        LIR & - & - & - & 37 & - \\
        ENR & - & - & - & 61 & - \\
        LAR & - & - & - & 58 & - \\
        DTR & - & - & - & 50 & - \\
        ABR & - & - & - & 24 & - \\
        GBR & - & - & - & 27 & - \\
        DYR & - & - & - & 36 & - \\ 
        XGBR & - & - & - & 22 & - \\
        RFR & - & - & - & 28 & - \\ 
        LSVR & - & - & - & 41 & - \\
        \bottomrule 
    \end{tabular}
    \caption{The mean combined task training and evaluation time per baseline model. Where applicable (cf. \Cref{appendix:baseline-performances}), the mean is computed across all model runs for various sections of the dataset.}
\end{table*}


\clearpage
\section{Expectations on Model Performance}
\label{appendix:expected-machine-performance}
We believe that our the baseline classifier performance can be vastly improved on and shifted towards the region of $0.8$ to $0.9$. We would expect tailored models to have near-perfect performance on the next part prediction task, RMSLEs of below $0.3$ (corresponding to $0.3$ uniform sequence element logarithm difference) in the continuation task, and top-$5$-RMSEs in the similarity and unmasking tasks of below $0.1$ and $1.0$, respectively.

Especially in the sequence similarity task, which is essentially a database lookup task, we would hope that the top-$k$-RMSE performance on the synthetic test set could be brought to essentially zero.
The OEIS dataset, however, contains many sequences that belong to the categories used for the experiments run in \Cref{section:baselines} but are very different from what one would achieve by procedural generation as per \Cref{appendix:synthetic}. We therefore do not think it feasible for the currently known architectures could achieve near-perfect results on this set, but see this as a natural challenge for architectures yet to be proposed.

An argument could be made that a human, if faced with the task of even just identifying the category of many of the sequences in OEIS, would struggle greatly and perhaps soon resort to guesswork.
The existence of Online Encyclopedia of Integer Sequences alone is a proof of that understanding integer sequences for what they are is, in practice, very often beyond the computational abilities of humans and demands thorough, encyclopedic knowledge instead.
By extension, one could be tempted to claim that our effort goes in the wrong direction with respect to the general motivation of artificial intelligence research, as the problem we are considering is much more difficult than what one would consider the natural baseline for intelligence -- human performance -- can reliably handle.
This dilemma of ``turning the tables in the Turing test'' was popularised and partially answered in \cite{penrose1990emperor}.
Let us just simply invoke the argument of the Turing test ``in the limit'', in which the human participant would have enough time and resources to develop a mechanism that would allow him to begin answering the trial's questions at computer-level speeds.
Then, the apparent difficulty of our benchmark is no longer in contrast with the common conception of intelligence, and furthermore, the development of such a mechanism is precisely something we hope will be facilitated by the benchmark presented in this paper.

The utility of our benchmark, however, can be seen even when disregarding this somewhat philosophical challenge.
The existence of OEIS as an online service proves demand for a system that ``understands'' sequences, and perhaps continues them, fills their gaps, and suggests similar sequences on top of immediately identifying which category the sequence belongs to.
Our benchmark can thus be viewed as a sort of ``CodeSearchNet \cite{husain2019codesearchnet} for sequences'', built to aid the development of models that can satisfy this demand.

\clearpage
\section{Relationship to Symbolic Regression}
\label{appendix:relationship-to-symbolic-regression}
One can notice that contemporary symbolic regression methods could be deemed suitable for performing the unmasking task, or at least its easier variant -- the continuation task.
We emphasise that the aim of the FACT toolkit is not to address these tasks as challenges, but instead to provide tools for research on the learning of finitary abstractions.

With this motivation in mind, there seems to be little evidence that classical symbolic regression methods manage to abstract any high-level descriptive information about the data they fit in their regression process.
Indeed, many of the methods used for symbolic regression perform systematic searches of hypothesis space of potential formulas.
While perhaps guided by sound empirical search methods and well-founded heuristics, they do not learn, and they do not learn to abstract anything about the nature of the sequences they regress.
Further, many of these methods make precisely the assumptions of simplicity of the underlying generative rule we make in our generative process, which would render their baseline evaluation results vacuous.
As such, we have therefore decided against their inclusion in this work, and believe that while state-of-the-art symbolic regression methods may play a role as performance contenders for future studies leveraging FACT, they would be difficult to interpret as natural baselines for finitary abstraction learning.


\clearpage
\section{An Illustrative Qualitative Evaluation of GPT-3 Davinci}\label{app:gpt3}
We qualitatively evaluate the performance of GPT-3 Davinci on initial segments of length 10 of the 6 most popular OEIS sequences for two tasks: sequence classification, and sequence continuation.
Our operating assumption is that the most popular sequences have been sufficiently present in the GPT-3 training data for the model to have a solid amount of information learned about them.
We have chosen temperature of $0.7$, maximum length of 256, top P of 1, frequency and presence penalty of $0$, and ``best of'' of 1.

\begin{dialogue}
\speak{Prompt} \direct{Continue the following sequence: } 1, 1, 2, 3, 5, 8, 13, 21, 34, 55
\speak{GPT-3 Davinci} 89
\speak{Baseline Transformer} 89
\speak{True (Assuming Fibonacci's sequence)} 89

\speak{Prompt} \direct{Continue the following sequence: } 0, 1, 3, 6, 2, 7, 13, 20, 12, 21
\speak{GPT-3 Davinci} 27, 28, 30, 33, 34, 36, 39, 40, 42, 45
\speak{Baseline Transformer} 11, 22, 10, 23, 9, 24, 8, 25, 43, 62
\speak{True (Assuming Recaman's sequence)} 11, 22, 10, 23, 9, 24, 8, 25, 43, 62

\speak{Prompt} \direct{Continue the following sequence: } 1, 1, 2, 5, 14, 42, 132, 429, 1430, 4862
\speak{GPT-3 Davinci} The next number in the sequence would be 16796.
\speak{Baseline Transformer} 16796
\speak{True (Assuming Catalan numbers)} 16796

\speak{Prompt} \direct{Continue the following sequence: } 2, 3, 5, 7, 11, 13, 17, 19, 23, 29
\speak{GPT-3 Davinci} 31, 37, 41, 43, 47, 53, 59, 61, 67, 71
\speak{Baseline Transformer} 31, 37, 41, 43, 47, 53, 59, 61, 67, 71
\speak{True (Assuming Prime numbers)} 31, 37, 41, 43, 47, 53, 59, 61, 67, 71

\speak{Prompt} \direct{Continue the following sequence: }  2, 3, 5, 7, 13, 17, 19, 31, 61, 89
\speak{GPT-3 Davinci} 131, 233, 337, 461, 563, 673, 761, 829, 937, 1033
\speak{Baseline Transformer} 107, 127, 521, 607, 1279, 2203, 2281, 3217, 4253, 4423
\speak{True (Assuming Mersenne exponents)} 107, 127, 521, 607, 1279, 2203, 2281, 3217, 4253, 4423

\speak{Prompt} \direct{Continue the following sequence: }  3, 7, 31, 127, 8191, 131071, 524287, 2147483647, 2305843009213693951, 618970019642690137449562111
\speak{GPT-3 Davinci} 618970019642690137449562111, 36893488147419103232, 1085102592571150095, 33554431, 
\speak{Baseline Transformer} \textit{Too big, the sequence was not included}
\speak{True (Assuming Mersenne primes)} 162259276829213363391578010288127, 170141183460469231731687303715884105727

\end{dialogue}

We observe that the GPT-3 sometimes performs well, although it takes liberties as to how it presents its output.
We further saw that our baseline transformer implementation for this task was consistently correct.

\begin{dialogue}
\speak{Prompt} \direct{Classify the following sequence: } 1, 1, 2, 3, 5, 8, 13, 21, 34, 55
\speak{GPT-3 Davinci} The sequence is an example of the Fibonacci sequence.
\speak{Baseline Transformer} \textit{N/A}
\speak{True (Assuming Fibonacci's sequence)} Fibonacci sequence.

\speak{Prompt} \direct{Continue the following sequence: } 0, 1, 3, 6, 2, 7, 13, 20, 12, 21
\speak{GPT-3 Davinci} The sequence is made up of numbers that follow the pattern of adding the two most recent numbers together. This pattern is known as a Fibonacci sequence.
\speak{Baseline Transformer} \textit{N/A}
\speak{True (Assuming Recaman's sequence)} Recamán's sequence.

\speak{Prompt} \direct{Continue the following sequence: } 1, 1, 2, 5, 14, 42, 132, 429, 1430, 4862
\speak{GPT-3 Davinci} The sequence appears to be the Fibonacci sequence.
\speak{Baseline Transformer} \textit{N/A}
\speak{True (Assuming Catalan numbers)} Catalan numbers.

\speak{Prompt} \direct{Continue the following sequence: } 2, 3, 5, 7, 11, 13, 17, 19, 23, 29
\speak{GPT-3 Davinci} 31, 37, 41, 43, 47, 53, 59, 61, 67, 71
\speak{Baseline Transformer Classifier} Prime.
\speak{True (Assuming Prime numbers)} Prime.

\speak{Prompt} \direct{Continue the following sequence: } 2, 3, 5, 7, 13, 17, 19, 31, 61, 89
\speak{GPT-3 Davinci} The sequence is a prime number sequence.
\speak{Baseline Transformer} Prime.
\speak{True (Assuming Mersenne exponents)} Prime.

\speak{Prompt} \direct{Continue the following sequence: } 3, 7, 31, 127, 8191, 131071, 524287, 2147483647, 2305843009213693951, 618970019642690137449562111
\speak{GPT-3 Davinci} The sequence appears to be a list of powers of two, with the exception of the third number.
\speak{Baseline Transformer} \textit{Too big, the sequence was not included}
\speak{True (Assuming Mersenne primes)} Prime, or rather, primes $p$ such that $2^p - 1$ is a prime.

\end{dialogue}

Here we observe that the GPT-3 is sometimes right, but sometimes just assumes a sequence based on its popularity, or misses crucial detail.


\clearpage
\section{Datasheets for Datasets}\label{app:datasheet}
For the dataset documentation we used the recommended
documentation framework "Datasheets for Datasets" \cite{gebru2018datasheets}.

\subsection{Motivation} \paragraph{For what purpose was the dataset created?}
To advance research in comprehension of finitary abstractions.

\paragraph{Who created the dataset?}
The dataset is a combination of the efforts of contributors to the Online Encyclopedia of Integer Sequences and our own systematic synthetic data generation effort. All those who contributed, directly and indirectly, are given credit in Acknowledgements.
We will aim to keep acknowledgements in our GitHub repository up-to-date as the work on the dataset continues in the future.

\paragraph{Who funded the creation of the dataset? If there is an associated
grant, please provide the name of the grantor and the grant name and
number}
\answerNA{}

\paragraph{Any other comments?} \answerNA{}

\subsection{Composition}
\paragraph{What do the instances that comprise the dataset represent (e.g.,
documents, photos, people, countries)? Are there multiple types of
instances (e.g., movies, users, and ratings; people and interactions between them; nodes and edges)?}
The dataset consists of integer sequences -- or rather, initial segments of integer sequences, and in most cases also annotations that allow for their automatic continuation if necessary.

\paragraph{How many instances are there in total (of each type, if appropriate)?}
See \Cref{section:dataset}. Short answer: About 3.6 million.

\paragraph{Does the dataset contain all possible instances or is it a sample
(not necessarily random) of instances from a larger set?}
It is a well-structured of a larger set as there are infinitely many integer sequences.

\paragraph{What data does each instance consist of? ``Raw'' data (e.g., unprocessed text or images) or features?}
Annotated entries pointing to integer sequences.

\paragraph{Is there a label or target associated with each instance? If so, please
provide a description}\answerYes{All labels and targets are described in \Cref{section:benchmark} and further \Cref{appendix:oeis}}.

\paragraph{Is any information missing from individual instances? If so, please
provide a description, explaining why this information is missing (e.g.,
because it was unavailable). This does not include intentionally removed
information, but might include, e.g., redacted text.}\answerNA{}

\paragraph{Are relationships between individual instances made explicit
(e.g., users’ movie ratings, social network links)? }  \answerNA{}

\paragraph{Are there recommended data splits (e.g., training, development/validation,
testing)?} \answerYes{ We have used 10:1:1 ratio of our synthetic training, synthetic testing, and organic testing sets in the benchmarking setup. Data split in the fashion is separately available through the ETH Research Collection, see \cite{fact_dataset_through_doi}.}

\paragraph{Are there any errors, sources of noise, or redundancies in the
dataset?} 
\answerNo{}

\paragraph{Is the dataset self-contained, or does it link to or otherwise rely on
external resources (e.g., websites, tweets, other datasets)?} The dataset is self-contained, there are no restrictions associated with any of the external resources that might apply to a future user.

\paragraph{Does the dataset contain data that might be considered confidential?}
\answerNo{}

\paragraph{Does the dataset contain data that, if viewed directly, might be offensive, insulting, threatening, or might otherwise cause anxiety?} \answerNo{}

\paragraph{Does the dataset relate to people?} \answerNo{}

\paragraph{Does the dataset identify any subpopulations (e.g., by age, gender)?} 
\answerNo{}

\paragraph{Is it possible to identify individuals (i.e., one or more natural persons), either directly or indirectly (i.e., in combination with other
data) from the dataset?} \answerNo{}
\paragraph{Does the dataset contain data that might be considered sensitive
in any way }
(e.g., data that reveals racial or ethnic origins, sexual
orientations, religious beliefs, political opinions or union memberships, or locations; financial or health data; biometric or genetic data; forms of government identification, such as social security numbers; criminal history)? 
\answerNo{}

\paragraph{Any other comments?} \answerNA{}

\subsection{Collection Process}
\paragraph{How was the data associated with each instance acquired? Was
the data directly observable (e.g., raw text, movie ratings), reported by
subjects (e.g., survey responses), or indirectly inferred/derived from other
data (e.g., part-of-speech tags, model-based guesses for age or language)?}
\answerYes{The collection and generation processes are extensively described in \Cref{section:benchmark}, \Cref{appendix:oeis}, \Cref{appendix:synthetic}.}

\paragraph{What mechanisms or procedures were used to collect the data
(e.g., hardware apparatus or sensor, manual human curation, software program, software API)? How were these mechanisms or procedures validated?}
\answerYes{The collection and generation processes are extensively described in \Cref{section:benchmark}, \Cref{appendix:oeis}, \Cref{appendix:synthetic}.}

\paragraph{If the dataset is a sample from a larger set, what was the sampling
strategy (e.g., deterministic, probabilistic with specific sampling
probabilities)?}
\answerYes{The collection and generation processes are extensively described in \Cref{section:benchmark}, \Cref{appendix:oeis}, \Cref{appendix:synthetic}.}

\paragraph{Who was involved in the data collection process (e.g., students,
crowdworkers, contractors) and how were they compensated (e.g.,
how much were crowdworkers paid)?}
\answerNA{}

\paragraph{Over what timeframe was the data collected?}
\answerNA{}

\paragraph{Were any ethical review processes conducted (e.g., by an institutional review board)?} 
\answerNA{}

\paragraph{Does the dataset relate to people?} \answerNo{}

\paragraph{ Did you collect the data from the individuals in question directly,
or obtain it via third parties or other sources (e.g., websites)?} \answerNA{}

\paragraph{Were the individuals in question notified about the data collection?} \answerNA{}
 
\paragraph{Did the individuals in question consent to the collection and use
of their data?} \answerNA{}

\paragraph{Has an analysis of the potential impact of the dataset and its use
on data subjects (e.g., a data protection impact analysis)been conducted?} \answerNA{}

\paragraph{Any other comments?}  \answerNA{}

\subsection{Preprocessing/cleaning/labeling}
\paragraph{Was any preprocessing/cleaning/labeling of the data done (e.g.,
discretization or bucketing, tokenization, part-of-speech tagging,
SIFT feature extraction, removal of instances, processing of missing values)?}
\answerYes{The collection, preprocessing, and generation processes are extensively described in \Cref{section:benchmark}, \Cref{appendix:oeis}, \Cref{appendix:synthetic}.}

\paragraph{Was the “raw” data saved in addition to the preprocessed/cleaned/labeled
data (e.g., to support unanticipated future uses)}

\paragraph{Is the software used to preprocess/clean/label the instances available? If so, please provide a link or other access point.} \answerYes{See the project code repository.}
\paragraph{Any other comments?} \answerNA{}

\subsection{Uses}
\paragraph{Has the dataset been used for any tasks already?} \answerYes{See \Cref{section:baselines}}. 
\paragraph{Is there a repository that links to any or all papers or systems that
use the dataset?} \answerNA{}

\paragraph{What (other) tasks could the dataset be used for?}
\answerYes{See \Cref{section:discussion}.}

\paragraph{Is there anything about the composition of the dataset or the way
it was collected and preprocessed/cleaned/labeled that might impact future uses?}
\answerNA{}

\paragraph{Are there tasks for which the dataset should not be used?} \answerNo{}
\paragraph{Any other comments?} \answerNA{}

\subsection{Distribution}
\paragraph{How will the dataset will be distributed (e.g., tarball on website,
API, GitHub)? Does the dataset have a digital object identifier (DOI)?}
The dataset is available through the ETH Research Collection (DOI 10.3929/ethz-b-000562705, \cite{fact_dataset_through_doi}).
The library (DOI 10.3929/ethz-b-000565638, \cite{factlib_through_doi}) and benchmarking models (DOI 10.3929/ethz-b-000565644, \cite{fact_dataset_through_doi}) are also available from the same source, and they too have their individual DOIs.

\paragraph{When will the dataset be distributed?}
The dataset is available through the ETH Research Collection, see \cite{fact_dataset_through_doi}.

\paragraph{Will the dataset be distributed under a copyright or other intellectual property (IP) license, and/or under applicable terms of use
(ToU)?} 
The dataset is distributed under the terms of the Creative Commons CC BY license, which permits non-commercial use, distribution, and reproduction in any medium, provided the original work is properly cited.

\paragraph{Have any third parties imposed IP-based or other restrictions on
the data associated with the instances?} \answerYes{Our license is in agreement with the license of the OEIS.}

\paragraph{Do any export controls or other regulatory restrictions apply to
the dataset or to individual instances?} \answerNo{}
\paragraph{Any other comments} \answerNA{}

\subsection{Maintenance}
\paragraph{Who is supporting/hosting/maintaining the dataset?}
The dataset is maintained by the FACT Development Team,  \url{https://github.com/FACT-Development-Team}.
In practice, this subsumes individual from the ETH D-ITET ISG and DISCO groups.
The hosting of the dataset is maintained by the ETH Research Collection \cite{fact_dataset_through_doi,factlib_through_doi,fact_benchmarking_baselines_through_doi}.

\paragraph{How can the owner/curator/manager of the dataset be contacted
(e.g., email address)?}
Email address: \href{mailto:fact@ethz.ch}{fact@ethz.ch}. Should this email address become unavailable in the future, please address your correspondence at members of ETH D-ITET DISCO.
\paragraph{Is there an erratum? If so, please provide a link or other access point} \answerNo{}
\paragraph{Will the dataset be updated (e.g., to correct labeling errors, add
new instances, delete instances)?} 
\answerYes{The dataset will be regularly updated, information about each update will be published in our repository.}

\paragraph{If the dataset relates to people, are there applicable limits on the
retention of the data associated with the instances (e.g., were individuals in question told that their data would be retained for a
fixed period of time and then deleted)?} \answerNA{}.

\paragraph{Will older versions of the dataset continue to be supported/hosted/maintained?}
\answerYes{Yes, all published versions of the dataset will be available through the ETH Research Collection.}

\paragraph{If others want to extend/augment/build on/contribute to the
dataset, is there a mechanism for them to do so? }
\answerYes{We release our complete infrastructure and provide a simple and easy-to-use interface to evaluate new methods. We invite the users of the dataset to submit their suggested contributions either directly to the authors or as pull requests through the GitHub system. The alternative is to contact us at \href{mailto:fact@ethz.ch}{fact@ethz.ch} to arrange closer collaboration if desired.}

\paragraph{Any other comments?} \answerNA{}

\section{Author Statement}
The authors of this work confirm that all parts of the content they publish conform to the licensing requirements of the individual parts, and that new additions are licenses accordingly. They further declare that they bear all responsibility in case of violation of any rights.

\end{document}